# AI–guided robotic in situ fabrication of curvature- and tissue-adaptive wireless bioelectronics

*Hiroyuki Tetsuka*, Jiaju Ma, Naomi Kumano, Minoru Hirano*

Bioinspired Systems Research-Domain, Toyota Central R&D Labs., Inc., 41-1 Yokomichi, Nagakute, Aichi 480-1192, Japan
E-mail: h-tetsuka@mosk.tytlabs.co.jp

**Abstract.** Personalized bioelectronic devices for wound therapy require not only conformal fabrication on complex body surfaces but also functional adaptation of their electrical characteristics to individual wound geometries. Here, we present an AI-guided robotic in situ printing platform for fabricating wireless wound electrotherapy devices directly on transparent wound dressings. The system integrates a 3D vision camera, a six-degree-of-freedom robotic arm, direct-ink-writing, and geometry-aware RF design optimization. AI-based wound segmentation is used to determine wound dimensions and local surface curvature, while the reconstructed three-dimensional geometry is incorporated into the electromagnetic design of the receiving coil and interdigitated electrodes. In contrast to conventional approaches that use surface geometry only for trajectory generation, the proposed method accounts for both anatomical curvature and tissue dielectric loading to compensate for geometry-induced shifts in resonant behavior. A printable silver ink was characterized in terms of rheology, dimensional reproducibility, electrical resistance, and electromechanical stability, and was used to fabricate multilayer wireless devices directly on curved polyurethane dressings. Wireless power transfer, impedance characteristics, and electric-field delivery were evaluated experimentally and numerically. Finally, robotic in situ fabrication was demonstrated on an anthropomorphic tissue-equivalent forearm phantom, where the measured dielectric properties of the phantom were incorporated into the RF optimization, enabling resonant operation near the target frequency of 25 MHz. This work establishes a perception-to-design-to-fabrication framework for geometry-adaptive wireless bioelectronics and provides a technological foundation for future physical-AI-enabled manufacturing of personalized on-body electrotherapy devices.

## 1. Introduction

Recent advances in robotics and additive manufacturing have enabled robot-assisted in situ fabrication, in which materials, biological constructs, and functional devices are directly fabricated on biological tissues or body surfaces while adapting to their local geometries[1-9]. In contrast to conventional manufacturing on planar substrates, in situ fabrication requires spatial information about the target, including its position, three-dimensional (3D) morphology, surface orientation, and curvature, to generate conformal printing trajectories. Adaptive printing systems incorporating computer vision have demonstrated direct fabrication of functional materials on moving freeform surfaces[1], while robot-assisted bioprinting has been applied to cartilage and bone regeneration[2,3], minimally invasive printing on curved tissues[4,5]. In addition, handheld in situ bioprinting[8] and intravital 3D bioprinting[9] have demonstrated the feasibility of directly fabricating biomaterials and tissue constructs in biological environments. These developments suggest that perception-guided robotic manufacturing could provide an important technological foundation for personalized biomedical fabrication, particularly in applications such as wound treatment, regenerative medicine, wearable electronics, and bioelectronic interfaces, where anatomical geometry varies substantially among individuals.

Multi-degree-of-freedom robotic manipulators offer particular advantages for such applications because they can control not only the 3D position of the printing nozzle but also its orientation relative to the target surface[2-6]. Compared with conventional three-axis printing systems, six-degree-of-freedom (6-DOF) robotic platforms can access highly curved anatomical surfaces and regions that cannot readily be approached from a single direction, including the lateral surfaces of limbs and areas surrounding joints[2,6] . When combined with 3D vision, these robotic systems can acquire geometrical information from the target and incorporate it into printing trajectories and tool poses. Such integration of perception and robotic motion enables fabrication that conforms to complex body surfaces rather than relying on predefined planar manufacturing coordinates[1,6,7].

Electrical stimulation represents a particularly attractive therapeutic function for integration with personalized in situ fabrication. Following disruption of the skin epithelium, endogenous electric fields arise around the wound and provide directional cues that regulate the migration of epithelial and other wound-associated cells[10]. Externally applied electric fields can mimic or reinforce these endogenous bioelectric signals and have been investigated for promoting cell migration, proliferation, re-epithelialization, tissue remodeling, and angiogenic responses during wound healing[10-13]. Accordingly, flexible and wearable electrotherapy platforms have emerged as promising alternatives to conventional wired stimulation systems[12-

[15]. Recent smart wound dressings have further incorporated wireless operation, sensing, and electrical stimulation, highlighting the potential of soft bioelectronics to provide localized therapy without restricting patient mobility[14,15].

Despite these advances, translating wireless electrotherapy from planar devices to anatomically conformal systems introduces an important challenge: the geometry of the body surface influences not only where a device should be fabricated, but also how the device functions electrically. Wound size, shape, position, and the curvature of the surrounding skin vary between individuals. These geometrical variations influence electrode placement and can also modify the electromagnetic characteristics of flexible receiving coils. Mechanical deformation of wireless antennas and resonators can alter their inductance, resonant frequency, impedance characteristics, and quality factor, thereby affecting wireless power-transfer performance[16]. Moreover, biological tissues provide frequency-dependent dielectric loading governed by their relative permittivity and electrical conductivity[17]. Consequently, transferring a device designed under planar, unloaded conditions directly onto a curved tissue surface may result in detuning from its intended operating frequency.

Personalized wireless wound bioelectronics have recently begun to address some of these challenges. In particular, the Customized wound patch for Advanced tissue Regeneration with Electric field (CARE) platform integrated a 6-DOF robotic arm, depth-camera-based wound measurement, customized interdigitated electrodes (IDEs), receiving-coil optimization, and wireless electrical stimulation[6]. The dimensions of the IDEs and receiving coil were customized according to the measured wound length and width, demonstrating the feasibility of rapidly manufacturing patient-specific wireless electrotherapy patches. However, in that approach, the electronic device was fabricated on a flexible substrate before being transferred to the target site. Although the RF design accounted for electrical component geometry and dielectric loading, the measured local surface curvature was not explicitly used to optimize the resonant characteristics of the device under its intended conformal configuration. Furthermore, fully automated transfer of sensed wound information to electronic-device optimization was identified as an important direction for further development. These considerations indicate that the next step toward anatomically adaptive wireless bioelectronics is to use the measured physical geometry not only to determine the printing path, but also to directly modify the functional electromagnetic design of the device.

Here, we propose an AI-guided robotic in situ printing system for personalized wireless wound electrotherapy devices directly fabricated on transparent wound dressings (Figure 1A). Rather than fabricating the electronic device on a planar substrate and subsequently transferring it to the wound, a transparent polyurethane dressing is first conformally applied over the wound, thereby providing both a protective wound barrier over the wound and a printable surface. The wound and surrounding anatomy are then characterized using 3D vision, after which a customized wireless electrotherapy device is computationally designed and directly fabricated onto the dressing. The resulting device consists of a printed receiving coil, dielectric layer, and IDEs and is designed to receive radio-frequency (RF) energy wirelessly from an external excitation coil. The proposed approach is primarily envisioned for superficial cutaneous wounds on anatomically curved body surfaces that can be covered with a conformal transparent dressing. Potential applications include localized bioelectronic stimulation of superficial skin injuries and ulcers, although therapeutic efficacy and applicability to different wound depths remain to be established.

The proposed system integrates a 3D vision camera, a 6-DOF robotic manipulator, a direct-ink-writing print head, and computational modules for AI-based image analysis, 3D geometrical reconstruction, RF design optimization, and robotic path planning (Figure 1B,C). First, an AI-based segmentation algorithm identifies the wound region from RGB images. The resulting segmentation mask is combined with the corresponding depth data to calculate the wound length, width, and projected area, while the local surface curvature is estimated from the three-dimensional point cloud surrounding the wound region. The reconstructed geometry is then used not only to define the conformal printing trajectory but also as an input to the electromagnetic design process. The geometry of the receiving coil and IDEs is optimized while accounting for both surface curvature and tissue-induced electromagnetic loading, thereby compensating for geometry-dependent changes in the resonant characteristics of the wireless device. The optimized device geometry is subsequently converted into a conformal printing path and corresponding robot poses for direct fabrication on the pre-applied transparent dressing.

This strategy extends conventional geometry-adaptive manufacturing beyond surface-following fabrication. Previous adaptive and robotic printing approaches have primarily used geometrical information to determine where and how a predefined structure should be printed[1-5,18]. In contrast, the present framework uses the measured wound geometry and local surface curvature not only to generate conformal printing trajectories but also to optimize the RF design of the receiving coil and interdigitated electrodes (IDEs). In this way, anatomical information is translated into both optimized electronic design parameters and robotic fabrication commands, enabling the direct fabrication of geometry-adaptive wireless bioelectronics. Such a perception–design–planning–fabrication framework represents a step toward physical AI and embodied intelligence by establishing a foundation for integrating perception, computational modeling, and robotic fabrication within a unified system[19,20]. Recent frameworks for physical AI similarly emphasize progression from perception and physical representation to reasoning, modeling, and embodied interaction with the environment.

In this study, we experimentally evaluate the proposed approach from perception through fabrication and functional validation. AI-based wound segmentation and geometrical characterization are assessed using wound phantoms with different shapes, skin appearances, and surface curvatures. The printable conductive ink is characterized in terms of rheological behavior, printing reproducibility, electrical properties, and electromechanical stability. Curvature- and tissue-aware RF optimization is evaluated by analyzing geometry-dependent changes in resonant characteristics, followed by characterization of wireless RF power transfer and electric-field delivery. Finally, the complete process is demonstrated on an anthropomorphic tissue-equivalent forearm phantom, in which experimentally determined dielectric properties are incorporated into the RF optimization and the resulting device is directly printed on a conformal transparent dressing and experimentally validated near the target operating frequency. Collectively, this work establishes a framework for geometry-adaptive manufacturing of personalized wireless bioelectronics, in which anatomical perception is directly coupled to both functional device design and robotic fabrication.

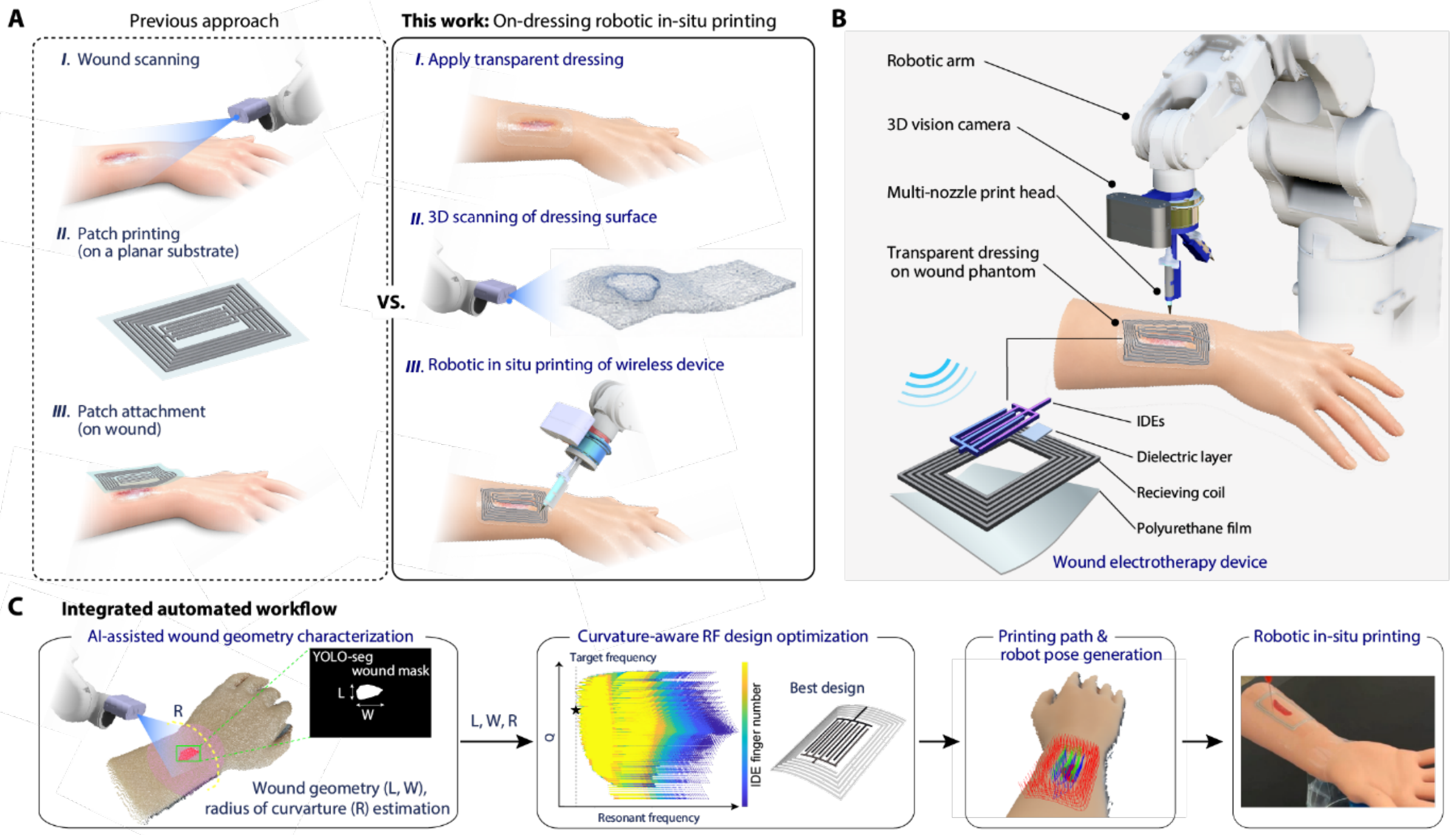

**Figure 1. Concept and workflow of robotic in situ printing of personalized wireless wound electrotherapy devices on transparent dressings.** (A) Comparison between the previously reported CARE patch-based approach and the proposed on-dressing robotic in situ printing

strategy. In the previous approach, the wound is scanned, a customized electrotherapy patch is fabricated on a planar substrate, and the patch is subsequently attached to the wound. In the proposed approach, a transparent dressing is first applied to the wound, followed by 3D scanning of the dressing surface and direct robotic printing of a personalized wireless electrotherapy device onto the dressing. (**B**) Schematic illustration of the robotic printing system. A robotic arm equipped with a 3D vision camera and a multi-nozzle print head performs direct printing of a multilayer wireless wound electrotherapy device on the transparent dressing. The printed device consists of a receiving coil, dielectric layer, and interdigitated electrodes (IDEs). (**C**) Integrated automated workflow of the proposed system. The workflow consists of AI-based wound segmentation from RGB images, three-dimensional surface reconstruction and geometric characterization using depth data, including wound dimension measurement and local curvature estimation, curvature- and tissue-aware RF design optimization of the receiving coil and interdigitated electrodes, printing-path and robot-pose generation, and robotic in situ printing of the personalized wireless wound electrotherapy device.

## 2. Results

### 2.1. Integrated AI-guided robotic workflow for personalized wireless bioelectronics fabrication

We first established an integrated AI-guided robotic manufacturing framework for the direct fabrication of personalized wireless electrotherapy devices on transparent wound dressings (Figure 1). In contrast to previous approaches in which a device is fabricated on a planar substrate and subsequently transferred to the treatment site, the proposed strategy uses a transparent polyurethane dressing that is first conformally applied to the target surface and then serves as the printing substrate (Figure 1A). The system integrates a six-degree-of-freedom robotic manipulator, a 3D vision camera, a direct-ink-writing print head, and computational modules for image analysis, RF design, and robotic path generation (Figure 1B). RGB and depth information acquired from the target surface were used to identify the wound region and characterize its geometry. The resulting geometrical parameters were automatically transferred to the RF design module, which generated a customized receiving coil and interdigitated electrode geometry. The optimized device geometry was subsequently converted into a three-dimensional printing trajectory and robot tool poses for direct fabrication on the transparent dressing (Figure 1C).

This integrated workflow enabled sequential execution of wound recognition, three-dimensional geometry estimation, RF design optimization, path planning, and robotic manufacturing without manual re-entry of the measured wound geometry into the device-design process. Thus, anatomical information obtained from the physical target was directly converted into both functional electronic design parameters and robotic fabrication commands. The total workflow time from AI-based wound geometry characterization to completion of four-pass robotic printing and drying was approximately 11 min 15 s (675 s). The computational stages required 40 s for geometry characterization, 15 s for MATLAB-based RF design optimization, and 10 s for printing-path generation, totaling 65 s. Robotic printing required 190 s across the four passes, while intermediate and final drying processes accounted for 420 s. These results demonstrate that the integrated geometry characterization, RF design, and printing-path generation processes could be completed in approximately 1 min under the investigated conditions, with drying accounting for the largest proportion of the total workflow time.

### 2.2. AI-based wound segmentation and local surface-curvature estimation

The ability of the vision system to identify wounds and quantify their geometry was evaluated using wound phantoms with different wound shapes, skin appearances, and surface geometries (Figure 2). YOLO-seg successfully extracted the wound regions from RGB images and generated pixel-level segmentation masks that remained consistent across the tested phantom conditions (Figure 2A,B). The segmented wound masks were registered with the corresponding depth information to determine wound length, width, and projected area. The estimated geometrical parameters showed close agreement with the predefined reference dimensions of the wound phantoms (Figure 2C), demonstrating that the AI-based segmentation could provide quantitative geometrical information required for subsequent device customization. The preparation of the wound-image dataset, training of the segmentation model, and representative real-time inference results are summarized in Supplementary Figure S1. To evaluate robustness under different lighting conditions, including variations in illumination color temperature, the AI-based method was compared with conventional color-based segmentation under ambient lighting and three controlled color-temperature conditions. The AI-based method detected all 16 wounds under each condition, whereas the color-based method detected 9 of 16 wounds under ambient lighting and only 2 of 16 under each controlled color-temperature condition (Supplementary Figure S2).

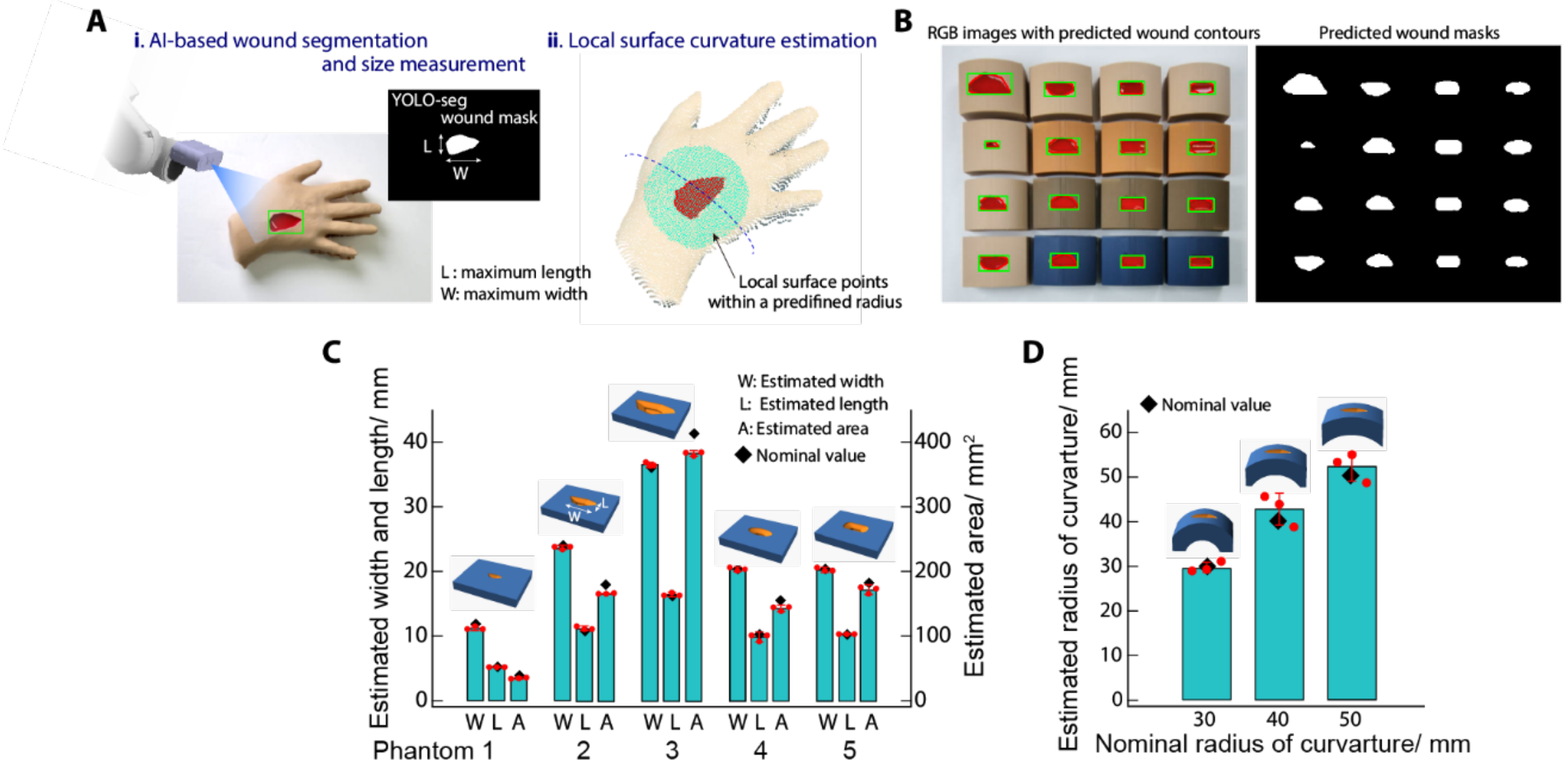

**Figure 2. AI-based wound segmentation, size measurement, and local surface curvature estimation.** (**A**) Workflow of AI-based wound segmentation and geometric characterization. Wound regions were segmented using YOLO-seg, and maximum length, maximum width, and projected area were calculated using the corresponding depth data. Local surface curvature was estimated from 3D points within a predefined radius of the wound centroid. (**B**) RGB images with predicted wound contours and the corresponding predicted binary masks. (**C**) Estimated maximum width, maximum length, and projected area for five wound phantom geometries. Black diamonds indicate nominal values, and red circles indicate individual measurements (n = 3 per phantom). (**D**) Estimated radii of curvature for phantoms with nominal radii of 30, 40, and 50 mm (n = 3 per phantom). Bars and error bars represent the mean and standard deviation, respectively. Measurement errors were calculated from individual measurements relative to the corresponding nominal values.

In addition to wound dimensions, local surface curvature was estimated from the three-dimensional point cloud surrounding the wound region. Curved phantoms with nominal radii of 30, 40, and 50 mm were used to evaluate the curvature-estimation procedure. The estimated radii followed the nominal curvature values across the tested geometries (Figure 2D), indicating that the vision system could capture not only two-dimensional wound dimensions but also the three-dimensional surface geometry required for conformal RF design and robotic printing. Quantitative analysis of five wound phantom geometries showed mean absolute errors (MAEs) of 0.45 mm and 0.39 mm for wound width and length, respectively, calculated from individual measurements (n = 3 per phantom). The projected wound area was systematically underestimated, with a mean absolute percentage error (MAPE) of 10.74% across 15 individual measurements from the five phantom geometries (Figure 2C). Local surface curvature estimation yielded an MAE of 2.57 mm and a MAPE of 6.16% across nine individual measurements obtained from phantoms with nominal radii of 30, 40, and 50 mm (Figure 2D). These results demonstrate that the vision-based approach can quantify wound dimensions and distinguish different surface curvatures, although systematic errors in area estimation and variability in curvature measurement remain.

### 2.3. Curvature- and tissue-aware RF optimization

We next investigated how anatomical curvature and tissue loading influence the RF characteristics of the wireless electrotherapy device (Figure 3). The device consisted of a receiving coil and IDEs integrated on a transparent polyurethane dressing, with an electromagnetic loading model incorporating the dressing layer and underlying tissue (Figure 3A). Candidate device geometries were generated by varying the IDE configuration and receiving-coil geometry. For each candidate design, the estimated resonant frequency and a quality-factor proxy incorporating conductor, substrate, and tissue-related losses were calculated. The optimal design was selected by minimizing a weighted objective function that prioritized resonant-frequency matching while favoring higher estimated quality factors and penalizing excessively small inner coil dimensions. This procedure provided a design space in which both electrode geometry and receiving-coil architecture could be selected according to the measured wound dimensions and local radius of curvature (Figure 3B,C).

Electromagnetic analysis further demonstrated that surface curvature influenced the resonant response of the wireless structure. To isolate the effect of curvature, the same dressing–skin electromagnetic loading model and material properties were used for both planar and curved configurations, with surface curvature as the only variable. For the same device geometry optimized for a curved surface with a radius of 45 mm, the simulated resonant frequency was 24.9970 MHz in the curved configuration, compared with 24.2803 MHz in the planar configuration (Figure 3D). This corresponds to a curvature-induced frequency shift of approximately 0.717 MHz (2.95% relative to the planar configuration). The optimized device exhibited a frequency deviation of only 0.012% from the target frequency of 25 MHz under the prescribed curved condition. These results highlight the importance of incorporating anatomical curvature into RF device design. Systematic variation of the radius of curvature similarly produced a curvature-dependent change in resonant frequency (Figure 3E). These results demonstrate that mapping a planar RF structure directly onto an anatomical surface does not necessarily preserve its original RF characteristics. Instead, surface curvature must be incorporated into the device-design process when resonance-dependent wireless operation is required.

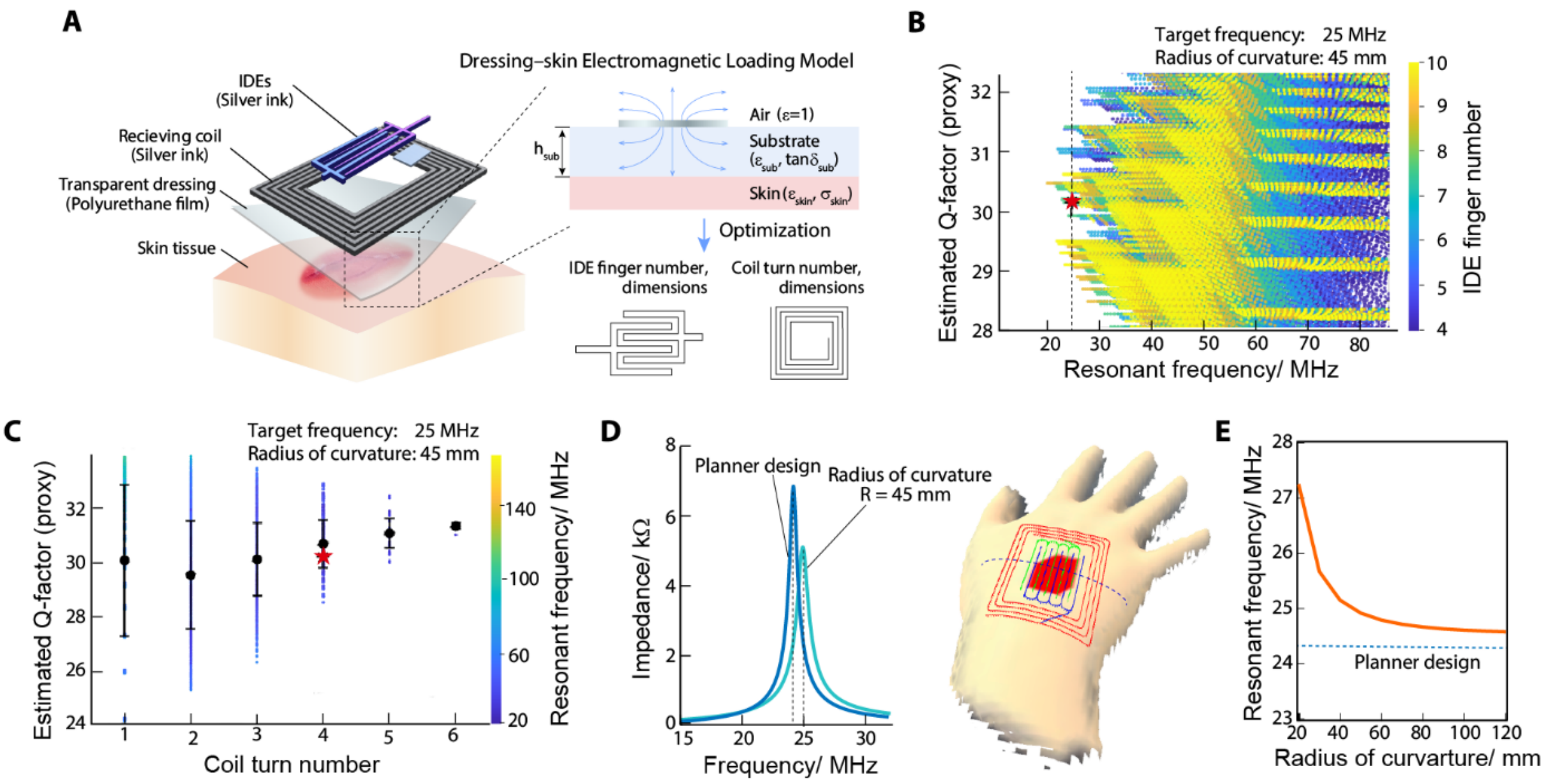

**Figure 3. Geometry-aware RF design optimization of the on-dressing wireless wound electrotherapy device.** (**A**) Exploded view of the device architecture and schematic of the dressing–tissue electromagnetic loading model used for RF design optimization. The model includes air, a transparent polyurethane dressing characterized by its relative permittivity, dielectric loss tangent, and thickness, and skin-equivalent tissue characterized by its relative permittivity and electrical conductivity. Candidate device geometries were generated by varying the receiving-coil and IDE design parameters. (**B**) Distribution of the estimated Q-factor proxy and resonant frequency for the candidate designs, color-coded according to the IDE finger number. The vertical dashed line indicates the target frequency of 25 MHz, and the star denotes the selected optimal design. (**C**) Dependence of the estimated Q-factor proxy on the receiving-coil turn number. Individual candidate designs are color-coded according to their estimated resonant frequencies. Black markers and error bars represent the mean and standard deviation of the Q-factor proxy across candidate designs for each coil turn number, respectively. The star indicates the selected optimal design. (**D**) Simulated impedance-magnitude spectra of the optimized device under planar and curved configurations using the same dressing–skin electromagnetic loading model, with surface curvature as the only variable. The inset shows the optimized device mapped onto the reconstructed hand-phantom surface with a radius of curvature of 45 mm. (**E**) Predicted resonant-frequency shift as a function of the radius of curvature. The horizontal dashed line indicates the resonant frequency under the planar configuration.

## 2.4. Rheological, printing, and electromechanical characteristics of the conductive ink

A printable silver ink containing silver flakes, silver nanoparticles, and PEO was developed for robotic direct writing on flexible polyurethane dressings (Figure 4A). Rheological measurements showed a pronounced reduction in apparent viscosity with increasing shear rate, indicating shear-thinning behavior suitable for extrusion-based printing (Figure 4B). The increasing and decreasing shear-rate sweeps exhibited similar overall trends, suggesting limited shear-history dependence under the tested conditions.

Strain-amplitude measurements further revealed a predominantly viscous viscoelastic response, with the loss modulus ($G''$) exceeding the storage modulus ($G'$) over most of the investigated strain range (Figure 4C). This behavior was consistent with an ink formulation designed for pressure-driven extrusion rather than solid-like shape retention prior to deposition. Continuous conductive traces could be directly printed on flexible polyurethane dressing substrates, including under curved configurations (Figure 4D). Under fixed printing conditions (24G nozzle, 20 mm $s^{-1}$ printing speed, 0.2 MPa pneumatic pressure, and 7% airflow controller setting), measurements of printed line width and thickness showed relatively consistent feature dimensions across three independently fabricated samples (Figure 4E,F).

Electrical resistance decreased markedly after printing and subsequently approached a lower steady-state value as the solvent evaporated and the conductive network developed (Figure 4G). Under repeated 50% compressive deformation, electrical continuity was maintained throughout 1000 loading cycles, although the relative resistance progressively increased during cyclic loading (Figure 4H). The effect of Ag filler composition on cyclic electromechanical stability was further evaluated by comparing the selected ink (85 wt% Ag flakes, 5 wt% Ag NPs, and 10 wt% PEO) with the Ag-flake-only formulation (90 wt% Ag flakes and 10 wt% PEO). The selected formulation exhibited a smaller increase in relative resistance ($R/R_0$) during compression cycles at 50% compressive strain (Supplementary Figure S3), indicating improved electrical stability under repeated deformation. Tensile deformation up to 50% similarly resulted in a gradual increase in normalized resistance while maintaining an electrically continuous conduction pathway (Figure 4I). These results indicate that the printed conductors tolerate substantial mechanical deformation while exhibiting strain-dependent changes in electrical resistance.

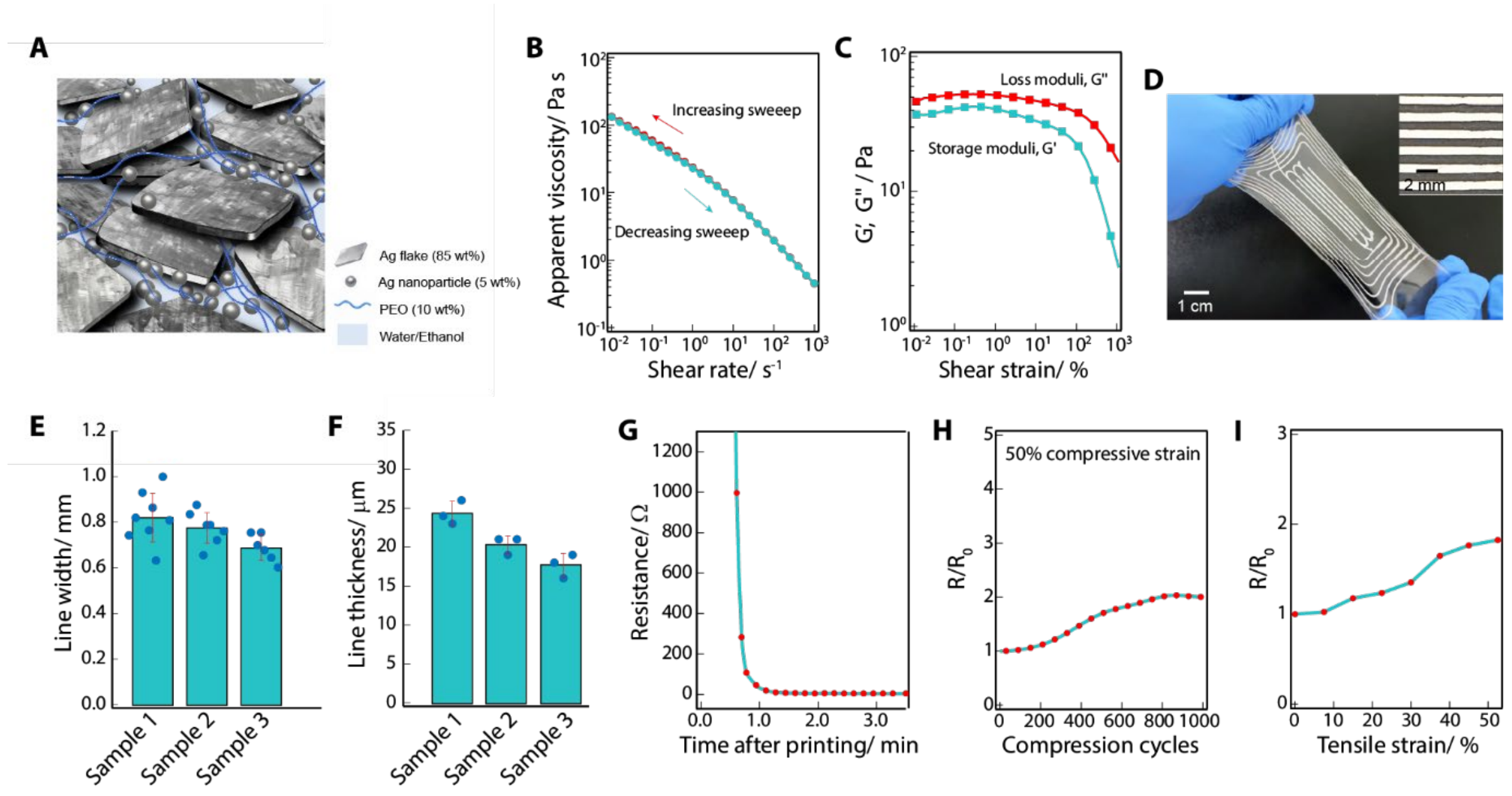

**Figure 4. Composition, rheological properties, printability, and electromechanical characterization of the printable silver ink.** (**A**) Schematic illustration of the silver ink formulation consisting of silver flakes, silver nanoparticles (nominal particle size <50 nm), polyethylene oxide (PEO; Mw ≈ 1,000,000 g mol$^{-1}$), and a water/ethanol solvent system. (**B**) Apparent viscosity of the silver ink as a function of shear rate during increasing and decreasing shear-rate sweeps from 0.01 to 1000 s$^{-1}$ at 25 °C. The decrease in viscosity with increasing shear rate indicates shear-thinning behavior. (**C**) Storage modulus (G′) and loss modulus (G″) as functions of shear strain, measured by a strain-amplitude sweep at 1 Hz and 25 °C. (**D**) Representative photograph of a conductive pattern printed on a flexible polyurethane dressing under bending deformation. The inset shows a magnified image of the printed lines. (**E,F**) Measured line width (E) and line thickness (F) for three independently printed samples fabricated under identical printing conditions using a 24G nozzle, a printing speed of 20 mm s$^{-1}$, an applied pneumatic pressure of 0.2 MPa, and an airflow controller setting of 7%. Bars and error bars represent the mean and standard deviation, respectively, and blue dots indicate individual measurements. (**G**) Change in the electrical resistance of the printed conductor as a function of post-printing time, showing rapid resistance reduction during solvent evaporation and drying. (**H**) Relative resistance (R/R0) of the printed conductor during 1000 compression cycles at 50% compressive strain. (**I**) Relative resistance of the printed conductor as a function of tensile strain up to 50%.

### 2.5. Wireless RF power transfer and electric-field delivery

The wireless operation of the printed device was subsequently evaluated using an external excitation coil (Figure 5A). A pulsed RF signal applied to the transmitter generated an induced voltage in the printed receiving coil, which in turn established an electric field between the IDEs. The wireless system was represented by a lumped-element equivalent circuit that included the transmitter and receiver coils, parasitic capacitance, IDE capacitance, dressing-mediated capacitive coupling, and the resistive and capacitive components of the tissue load (Figure 5B). This model illustrates that the RF behavior of the printed device is determined by both the receiving structure itself and the electromagnetic loading introduced by the dressing and underlying tissue.

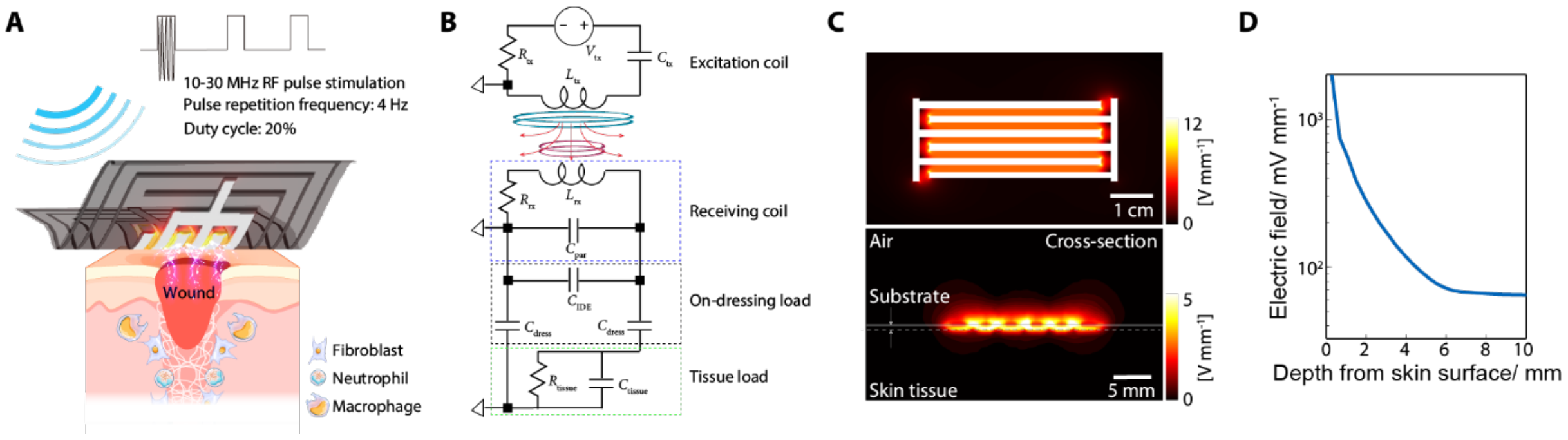

**Figure 5. Wireless RF power transfer and electric-field delivery of the on-dressing wound electrotherapy device.** (**A**) Schematic illustration of the wireless wound electrotherapy system. A pulsed RF signal is wirelessly transmitted from an external excitation coil to the printed receiving coil, generating an electric field across the IDEs for wound stimulation. (**B**) Equivalent circuit model of the wireless power transfer system, including the excitation coil, receiving coil, IDE capacitance, polyurethane dressing coupling capacitance, and tissue loading impedance. (**C**) Simulated electric-field distribution generated by the IDEs. The upper panel shows the surface electric-field distribution, while the lower panel presents the cross-sectional electric-field distribution through the polyurethane dressing and skin tissue. (**D**) Simulated electric-field magnitude as a function of depth from the skin surface, demonstrating penetration of the electric field into the underlying tissue.

Electromagnetic simulations predicted a spatially distributed electric field generated by the IDE structure over the intended treatment region (Figure 5C). Cross-sectional analysis further indicated that the simulated field extended through the transparent dressing and into the underlying tissue-equivalent domain. The electric-field magnitude decreased with increasing depth from the tissue surface, reflecting the combined effects of geometrical spreading, dielectric properties, and conductive losses (Figure 5D).

In our previous study, an electric-field magnitude of 25 mV mm$^{-1}$ was discussed as a reference level associated with electrically guided cell migration[6]. However, this value was not used as a prescribed therapeutic threshold or optimization target in the present study, because the effective stimulation conditions for pulsed RF excitation have not yet been established. Instead, the present simulations were intended to characterize the spatial distribution and depth-dependent variation of the electric field within the tissue-equivalent model. At a depth of 5 mm from the tissue surface, the simulated maximum and mean electric-field magnitudes were 83.9 and 44.0 mV mm$^{-1}$, respectively. At a depth of 10 mm, the corresponding values were 64.8 and 32.7 mV mm$^{-1}$. These results indicate that the simulated electric field extended into the tissue-equivalent domain under the prescribed excitation and boundary conditions. The evaluated depths represent locations for characterizing electric-field distribution rather than established therapeutic target depths. Together, the experimental wireless RF reception measurements and numerical electric-field simulations support the feasibility of integrating wireless signal reception with electric-field delivery across a dielectric wound-dressing interface without a direct wired connection.

### 2.6. Robotic in situ fabrication and RF validation on an anthropomorphic tissue-equivalent forearm phantom

For experimental validation on the anthropomorphic forearm phantom, the wireless device was optimized to operate near the target frequency of 25 MHz by incorporating the measured dielectric properties and local surface curvature of the tissue-equivalent phantom into the RF design process. The transparent polyurethane dressing conformed to the curved phantom surface and provided a stable substrate for robotic direct printing (Supplementary Figure S4). To reproduce the electromagnetic environment of the target substrate, the experimentally determined dielectric properties of the tissue-equivalent phantom at 25 MHz were incorporated into the RF model. A relative permittivity of $\varepsilon_r = 125$ and conductivity of $\sigma = 0.31\ \mathrm{S\ m^{-1}}$were used together with the measured surface curvature for tissue- and curvature-aware device optimization (Figure 6A).

Numerical analysis using the dry-skin-equivalent loading model predicted a curvature-dependent change in the frequency response of the optimized device (Figure 6B). The simulated resonant frequency shifted from 23.87 MHz in the planar configuration to 25.16 MHz in the curved configuration corresponding to the anthropomorphic forearm phantom, representing a frequency shift of 1.29 MHz (5.40% relative to the planar configuration). This numerical comparison indicates that conformal curvature can substantially influence the resonant characteristics of the wireless device, highlighting the importance of incorporating anatomical geometry into RF design optimization.

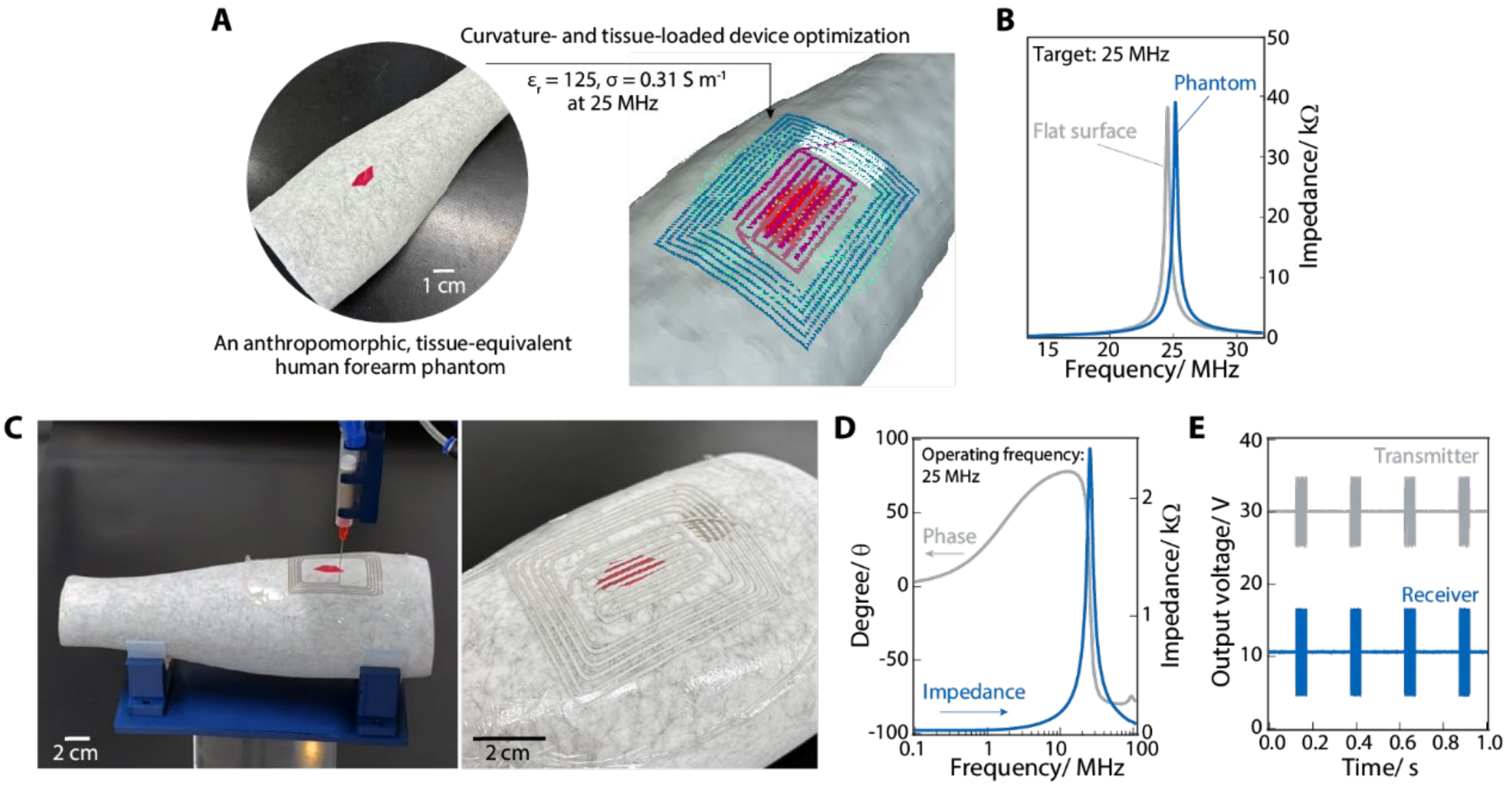

**Figure 6. Robotic in situ fabrication and experimental RF validation of the optimized wireless electrotherapy device on an anthropomorphic tissue-equivalent forearm phantom.** (**A**) Photograph of the anthropomorphic tissue-equivalent forearm phantom and curvature- and tissue-aware RF optimization of the device using the dielectric properties of the phantom measured at 25 MHz. (**B**) Simulated frequency-dependent impedance responses of the optimized device under planar and curved configurations with dry-skin-equivalent dielectric loading, illustrating the predicted influence of surface curvature on the resonant characteristics. (**C**) Robotic in situ printing of the optimized wireless electrotherapy device directly onto a transparent dressing conformally attached to the tissue-equivalent forearm phantom and representative photograph of the resulting device on the curved phantom surface. (**D**) Measured impedance magnitude and phase spectra of the optimized device fabricated on the tissue-equivalent forearm phantom, demonstrating resonant operation near the target frequency of 25 MHz. **(E)** Representative transmitter and receiver voltage waveforms measured during wireless operation on the tissue-equivalent forearm phantom, confirming successful wireless RF signal transfer.

The optimized device was subsequently fabricated directly on the transparent dressing attached to the tissue-equivalent forearm phantom using the six-axis robotic system (Figure 6C). Experimental impedance and phase measurements revealed a resonant frequency of 25.68 MHz (Figure 6D), corresponding to a deviation of only 0.52 MHz (2.07%) from the simulated phantom-loaded resonance of 25.16 MHz. This agreement supports the predictive capability of the curvature- and tissue-aware RF design approach under the investigated phantom conditions.

Wireless measurements further confirmed successful reception of the pulsed RF signal by the printed device on the phantom (Figure 6E). Additional measurements using two different transmitter configurations demonstrated wireless signal reception with the transmitter positioned either beneath or above the forearm phantom (Supplementary Figure S5). Together, these results demonstrate that anatomical curvature and tissue-equivalent electromagnetic loading can be incorporated into the RF design process before robotic fabrication, enabling the fabrication of wireless bioelectronic devices with resonant characteristics close to the prescribed operating frequency under anatomically conformal conditions.

## 3. Discussion

The present study demonstrates a robotic manufacturing framework in which anatomical perception is directly linked to the functional design and physical fabrication of wireless bioelectronics. The central distinction of this approach is that three-dimensional geometrical information is not used solely for generating a surface-following printing trajectory. Instead, the measured wound geometry and surface curvature are also incorporated into the RF design of the receiving coil and electrodes before fabrication. Thus, the same sensory information influences both where the device is fabricated and how its functional geometry is designed.

Previous robotic in situ printing studies have demonstrated capabilities for depositing biomaterials or functional inks on planar, curved, moving, and internally accessible biological surfaces[1-5,18]. More recently, Yoo et al. demonstrated robotic conformal printing of multilayer electronic displays on complex three-dimensional surfaces using a six-degree-of-freedom robotic arm and three-dimensional scanning [21]. These studies established the importance of three-dimensional sensing and multi-degree-of-freedom robotic control for conformal fabrication. However, these approaches primarily focus on adapting predefined functional patterns and printing trajectories to the target geometry. In contrast, the present framework incorporates the measured surface curvature and tissue-dependent electromagnetic loading into the functional RF design itself, enabling both device geometry and robotic fabrication trajectories to be adapted to the target anatomical environment. This distinction is particularly important for wireless electronics, because conformal deformation can alter the electrical response of a device rather than merely its spatial position.

Our RF simulations indicate that surface curvature influences the resonant behavior of the receiving structure, while experimental measurements on the anthropomorphic tissue-equivalent forearm phantom support the predictive capability of the curvature- and tissue-aware RF design approach. Such behavior is consistent with previous observations that deformation of flexible wireless antennas and resonators alters inductive, capacitive, and impedance characteristics[16]. In the present system, this problem is compounded by electromagnetic loading from the underlying biological tissue, whose relative permittivity and conductivity strongly influence near-field RF behavior[17]. Consequently, a wireless structure optimized under planar and unloaded conditions may not retain the same resonant frequency after being placed on a tissue-like surface.

The use of curvature- and tissue-aware optimization therefore represents an important step beyond purely geometrical path adaptation. Particularly for resonantly coupled wireless devices, even moderate shifts in resonance can reduce received voltage and power-transfer performance. By incorporating local surface curvature and tissue dielectric properties into the RF model before fabrication, the proposed framework allows the device geometry to be adjusted for its intended deployment environment.

The anthropomorphic forearm-phantom experiments provide practical support for this concept. Numerical simulations predicted that the resonant frequency of the optimized device shifted from 23.87 MHz in the planar configuration to 25.16 MHz under the curved, tissue-equivalent loading condition, corresponding to a frequency shift of 1.29 MHz (5.40% relative to the planar configuration). Importantly, after incorporating the measured tissue-equivalent dielectric properties and anatomical curvature into the RF optimization, the device was directly fabricated on the transparent dressing attached to the anthropomorphic forearm phantom. The experimentally measured resonant frequency was 25.68 MHz, differing from the simulated phantom-loaded value by 0.52 MHz (2.07%). This agreement between numerical prediction and experimental measurement supports the feasibility of curvature- and tissue-aware RF design for robotic in situ fabrication. Together, these results establish a quantitative connection between the numerical design optimization in Figure 3 and the experimental validation in Figure 6.

The proposed on-dressing fabrication strategy also differs from previously reported CARE approaches. The CARE platform demonstrated robot-assisted wound measurement, customized electrode and receiving-coil design, wireless power transfer, and electrical stimulation[6]. That work established an important foundation for rapidly customized wireless wound bioelectronics. Although the CARE design considered dielectric loading, device customization was primarily based on wound dimensions, and the measured local surface curvature was not explicitly incorporated into the RF optimization. Furthermore, fabrication was performed on a flexible substrate prior to placement at the wound site. In contrast, the present approach explicitly incorporates the measured local surface curvature into the RF design optimization and directly fabricates the optimized device on a transparent dressing already conformally attached to the target surface. This strategy can reduce the geometrical discrepancy between the fabrication state and the final deployed state while accounting for curvature-dependent changes in the resonant characteristics of the wireless device. The potential advantages of robotic on-dressing fabrication are particularly relevant to anatomically complex regions, such as the dorsal hand, forearm, and areas surrounding joints, where local surface curvature and orientation can vary substantially. Although conventional flexible wound patches can conform mechanically to curved body surfaces, deformation during attachment may alter their intended geometry and resonant characteristics. In contrast, the proposed approach characterizes the target geometry after the transparent dressing has been applied and incorporates the measured curvature into both RF design optimization and robotic printing trajectories. This strategy may reduce discrepancies between the designed and deployed device configurations and facilitate individualized electrode placement without transferring a separately fabricated device onto the wound. Nevertheless, robotic in situ fabrication requires additional sensing, dispensing, and motion-control infrastructure compared with conventional patch manufacturing. Its practical advantages are therefore expected to be most relevant to anatomically complex surfaces requiring geometry-specific device customization, rather than applications in which standardized flexible patches are sufficient.

The transparent dressing also provides a useful functional interface between the printed electronics and the wound. Direct deposition of conductive ink onto exposed tissue would introduce concerns related to material compatibility, electrical contact, removal, and contamination. By printing onto a pre-applied dressing, the proposed approach physically separates the conductive materials from the wound while preserving the possibility of delivering an electric field through capacitive coupling. The dressing therefore simultaneously serves as a wound-protective layer, a mechanically compliant printing substrate, and part of the electrical interface.

From an application perspective, this on-dressing fabrication strategy is primarily intended for superficial cutaneous wounds, including shallow skin injuries and ulcers, on anatomically curved body surfaces. Its applicability is expected to depend on wound depth, exudate levels, and the suitability of the transparent dressing for the specific wound condition.

Further studies are required to evaluate the biological response, safety, and therapeutic efficacy of the printed device in clinically relevant wound models.

Although electromagnetic simulations predicted electric-field distributions at depths of 5 and 10 mm within the tissue-equivalent model (Figure 5D), these depths were selected to characterize the spatial distribution of the electric field rather than to define specific therapeutic targets. The effective electric-field magnitude and stimulation depth required for wound healing under pulsed RF excitation remain to be established. The previously reported CARE platform also investigated electric-field delivery into biological tissues; therefore, the present results should not be interpreted as demonstrating superior electric-field penetration or therapeutic efficacy compared with the CARE approach. Instead, the principal contribution of this study lies in incorporating measured local surface curvature and tissue dielectric properties into RF design optimization, followed by direct robotic fabrication on a conformal wound dressing. Further biological studies using clinically relevant wound models will be required to determine appropriate stimulation conditions for different wound types and depths.

The rheological and electromechanical results further support the feasibility of this manufacturing approach. The silver ink exhibited shear-thinning behavior suitable for extrusion-based deposition, while continuous conductive patterns could be formed on flexible and curved polyurethane substrates. The ink formulation was based on the Ag flake/PEO system reported by Zhu et al.[1], while maintaining a total Ag-to-PEO weight ratio of 9:1 in the dried ink. Partial replacement of Ag flakes with Ag nanoparticles (Ag NPs) improved the cyclic electrical stability, with the 85 wt% Ag flake/5 wt% Ag NP/10 wt% PEO formulation exhibiting a smaller increase in $R/R_0$ during repeated compression than the 90 wt% Ag flake/10 wt% PEO formulation (Supplementary Figure S3). The improved cyclic electrical stability may result from additional conductive pathways formed by Ag NPs between adjacent Ag flakes, although the underlying mechanism requires further investigation[22 23]. These results suggest that incorporating a small fraction of Ag NPs into the Ag flake/PEO formulation can improve electrical stability under cyclic deformation[22 23 24].

The integrated computational workflow is another important aspect of the system. Previous personalized wound-electronics platforms have identified manual transfer between wound sensing and electronic optimization as a limitation to full automation. In the present framework, AI-based wound segmentation, depth-based three-dimensional geometrical analysis, RF design optimization, and robotic path generation are sequentially integrated under Python control, with the MATLAB-based RF optimization routine executed automatically. This architecture enables automated transfer of geometrical information between perception, device design, and robotic fabrication, reducing the need for manual data transfer between computational stages. Although preparatory operations, such as dressing placement and ink loading, remain manual, the proposed system establishes an integrated perception–design–planning–fabrication workflow.

The integrated workflow was completed in approximately 11 min, including geometry characterization, RF design optimization, robotic printing, and drying. Notably, the computational stages required 65 s, whereas drying accounted for approximately 62% of the total workflow time. These findings suggest that further improvements in drying processes could substantially reduce the overall fabrication time.

From a broader perspective, this workflow can be interpreted within the emerging framework of physical AI[19,20]. The system perceives its physical environment through RGB-depth sensing, extracts task-relevant geometrical information using AI, converts that information into a functional electronic design, generates robot actions, and physically modifies the environment through manufacturing. The role of AI-based perception therefore extends beyond image classification: its outputs are directly coupled to physical design and action. This type of integration may be particularly valuable for personalized manufacturing, where the optimal manufactured structure cannot be completely predefined before the physical target is observed.

Several limitations remain. First, robotic on-dressing fabrication and RF performance were validated using tissue-equivalent phantoms, but the biological performance of devices directly fabricated on wound dressings has not yet been evaluated in living tissue. Biological surfaces exhibit additional complexity, including motion, hydration, spatially heterogeneous dielectric properties, wound exudate, and time-dependent deformation. Second, although surface curvature and bulk tissue loading were incorporated into the RF optimization, future models should account for multilayer skin structures, wound fluid, dressing hydration, and spatial variation in dielectric properties. Third, the current system uses an open-loop fabrication process with respect to target-surface motion and deformation after geometry acquisition. Real anatomical targets may move during printing, requiring real-time visual tracking or force feedback to compensate for motion and deformation. Fourth, long-term electrical and mechanical stability of the printed device under repeated body motion, perspiration, moisture exposure, and dressing replacement must be further evaluated.

Future versions of the system could integrate printed temperature, impedance, and pH sensors with the stimulation electrodes. Such multimodal sensing could enable continuous monitoring of the wound environment and adaptive adjustment of stimulation parameters.

Combined with real-time perception and closed-loop robotic control, this would extend the present perception–design–fabrication framework toward a closed-loop physical-AI platform capable of sensing, fabricating, monitoring, and adapting treatment according to changes in the wound environment.

## 4. Conclusion

We developed an AI-guided robotic in situ manufacturing platform for the personalized fabrication of wireless wound electrotherapy devices directly on transparent dressings. The system integrates AI-based wound segmentation, three-dimensional surface characterization, curvature- and tissue-aware RF optimization, robotic path and pose generation, and direct-ink-writing fabrication within a continuous computational workflow. The proposed approach differs from previously reported geometry-adaptive printing by using the measured anatomical geometry not only to generate the printing trajectory but also to determine the functional RF design of the device. Surface-curvature analysis demonstrated that conformal deformation can alter resonant characteristics, while experimental measurements on a tissue-equivalent phantom showed good agreement with the predicted resonant frequency under combined curvature and electromagnetic loading conditions.

By incorporating both surface curvature and tissue dielectric properties into the design process, the optimized device could be directly fabricated on an anthropomorphic tissue-equivalent forearm phantom and exhibited a resonant response near the target operating frequency of 25 MHz. The printable silver ink provided shear-thinning behavior suitable for robotic extrusion and maintained conductive pathways under substantial mechanical deformation. Wireless measurements confirmed RF signal reception by the printed device, while electromagnetic simulations predicted electric-field delivery through the transparent dressing interface.

Collectively, these results establish a geometry- and tissue-aware robotic manufacturing framework for personalized wireless bioelectronics. More broadly, the study demonstrates a physical-AI-inspired approach in which perception of the physical environment is translated directly into functional device design, robotic action planning, and physical fabrication. This concept may provide a foundation for future personalized bioelectronic systems that are designed and manufactured directly according to patient-specific anatomical and physiological conditions.

## Materials and Methods

*Integrated AI–robotic system for in situ device fabrication*

An integrated AI-guided robotic manufacturing system was developed to automate the entire workflow from wound recognition and geometrical characterization to RF device design, printing-path generation, robot-pose planning, and direct fabrication (Figure 1C). The system consisted of a six-degree-of-freedom industrial robotic manipulator (C8-A701S, Epson, Japan), a 3D vision camera (Femto Bolt, Orbbec, China), a custom-built direct-ink-writing print head, and a host computer for image processing, design optimization, and robot control. An eye-in-hand calibration procedure using a ChArUco target was performed to determine the transformation between the RGB-D camera and robot tool coordinate systems. The calibration procedure, transformation equations, quality filtering, and algorithm selection are described in Supplementary Text S1.

Python was used as the supervisory control environment to integrate the individual computational and robotic processes. RGB and depth data acquired using the 3D vision camera were processed to identify the wound region and determine the maximum wound length ($L$), maximum width ($W$), projected area ($A$), and local radius of curvature ($R$) of the surrounding surface. The extracted geometrical parameters were subsequently transferred to an RF design module implemented in MATLAB (MathWorks, USA).

The MATLAB-based optimization program was executed automatically from Python in batch mode. The wound dimensions and local surface curvature were used as input parameters for optimization of the receiving coil and interdigitated electrodes (IDEs). Upon completion of the RF optimization, the selected device geometry was automatically returned to the Python environment. The resulting two-dimensional device geometry was then mapped onto the reconstructed three-dimensional surface to generate conformal printing trajectories and corresponding robot tool poses. The optimized device geometry was projected onto the reconstructed three-dimensional surface, and the corresponding robotic tool positions and orientations were generated using local surface normals and trajectory directions. Details of surface projection, tool-orientation smoothing, trajectory resampling, and robotic path execution are provided in Supplementary Text S2.

The generated position and orientation information was used to control the Epson C8-A701S robotic manipulator and execute direct printing onto a transparent wound dressing. Thus, the entire workflow—AI-based wound recognition, 3D geometry and curvature estimation, wireless-device design optimization, printing-path and robot-pose generation, and robotic

fabrication—was performed sequentially under integrated Python control without manual modification of the device design parameters after wound characterization.

*AI-based wound segmentation and three-dimensional geometrical characterization*

RGB and depth images of the wound phantoms were acquired using the Femto Bolt 3D vision camera (Orbbec) to characterize the wound geometry and surrounding surface curvature (Figure 2A). Wound regions were identified using YOLO-seg, an instance-segmentation model capable of pixel-level object segmentation. The trained model was applied to RGB images to generate binary masks of the wound regions. The masks were registered with the corresponding depth images, and the resulting spatial information was used to calculate the maximum wound length ($L$), maximum width ($W$), and projected wound area ($A$). To evaluate the robustness of the segmentation procedure, wound phantoms with different wound geometries and skin appearances were analyzed. The calculated geometrical parameters were compared with the predefined reference dimensions of the corresponding phantom wounds.

To evaluate measurement accuracy, each wound phantom was measured three times. The mean absolute error (MAE) was calculated by averaging the absolute differences between individual measurements and their corresponding nominal values across all tested phantoms. The mean absolute percentage error (MAPE) was calculated by averaging the absolute errors normalized by the corresponding nominal values and expressed as percentages. This calculation procedure was consistently applied to wound width, length, projected area, and surface radius of curvature.

*Curvature- and tissue-aware RF design of the wireless electrotherapy device*

The wireless wound electrotherapy device consisted of a printed receiving coil, dielectric layer, and interdigitated electrodes fabricated on a transparent polyurethane dressing (Figure 3A). To account for the electromagnetic environment encountered during conformal placement on the body, an RF loading model incorporating air, the polyurethane dressing, and skin-equivalent tissue was constructed. The polyurethane dressing was modeled using its relative permittivity, dielectric loss tangent, and thickness, whereas the underlying tissue was represented by its relative permittivity ($\varepsilon_r$) and electrical conductivity ($\sigma$). These parameters were incorporated into the RF design model to approximate dielectric and conductive loading on the printed wireless device.

The design variables included the number of receiving-coil turns, coil dimensions, number of IDE fingers, and IDE dimensions. The wound length ($L$), width ($W$), and local radius of curvature ($R$) obtained from the 3D vision system were automatically passed to the MATLAB-based optimization routine. For each candidate geometry, the resonant frequency was estimated as

$$f_{\mathrm{est}} = \frac{1}{2\pi\sqrt{L_c C_{\mathrm{IDE}}}},$$

where $L_c$ and $C_{\mathrm{IDE}}$ represent the curvature-corrected coil inductance and effective IDE capacitance, respectively.

To evaluate RF performance under tissue-loading conditions, an approximate quality-factor metric was calculated by combining conductor and dielectric losses:

$$Q_{\mathrm{proxy}} = \left(\frac{1}{Q_{\mathrm{coil}}} + \frac{1}{Q_{\mathrm{dielectric}}}\right)^{-1},$$

where

$$Q_{\mathrm{coil}} = \frac{2\pi f_0 L_c}{R_{\mathrm{dc}}}, \qquad Q_{\mathrm{dielectric}} = \frac{1}{\tan\delta_{\mathrm{eff}}}.$$

Here, $f_0$ is the target operating frequency, $R_{\mathrm{dc}}$ is the estimated coil resistance calculated from the conductor length, cross-sectional area, and electrical conductivity, and $\tan\delta_{\mathrm{eff}}$ is the effective loss tangent incorporating substrate dielectric loss and approximate tissue-induced conductive loss. The resulting quality-factor metric was used as a simplified indicator of RF losses rather than a rigorously calculated loaded quality factor.

Candidate designs were ranked by minimizing a weighted objective function:

$$S = \alpha_f \frac{|f_{\mathrm{est}} - f_0|}{f_0} + \frac{\alpha_Q}{Q_{\mathrm{proxy}} + 10^{-6}} + \alpha_{\mathrm{in}} P_{\mathrm{in}},$$

where $\alpha_f = 1.0$, $\alpha_Q = 0.05$, and $\alpha_{\mathrm{in}} = 0.10$. The penalty term $P_{\mathrm{in}}$ was introduced to discourage excessively small inner coil openings and was defined as

$$P_{\mathrm{in}} = \max\left(0, \frac{L_{\mathrm{soft}}}{L_{\mathrm{in}}} - 1\right) + \max\left(0, \frac{W_{\mathrm{soft}}}{W_{\mathrm{in}}} - 1\right),$$

where $L_{\mathrm{in}}$ and $W_{\mathrm{in}}$ are the inner coil dimensions, and $L_{\mathrm{soft}}$ and $W_{\mathrm{soft}}$ are the soft minimum dimensions.

The candidate with the lowest objective score was selected for fabrication. This optimization strategy prioritized resonant-frequency matching while favoring higher estimated quality factors and penalizing excessively small inner coil openings. To quantify the effect of conformal deformation on the RF response, the frequency-dependent impedance magnitude of the same device geometry was also evaluated under planar and curved configurations. In addition, the radius of curvature was systematically varied to determine the relationship between surface curvature and resonant frequency. To isolate the influence of surface curvature on the resonant response, the optimized device geometry was evaluated under planar and curved configurations using identical dressing–skin dielectric properties, material thicknesses, and tissue-loading conditions, with surface curvature as the only variable.

*Preparation and rheological characterization of the conductive silver ink*

A printable silver-based conductive ink was prepared for robotic direct writing. Silver flakes (product no. 327077), silver nanoparticles (nominal particle size <50 nm; product no. 730807), and polyethylene oxide (PEO; weight-average molecular weight, Mw ≈ 1,000,000 g $mol^{-1}$; product no. 372781) were obtained from Sigma-Aldrich. The silver flakes and nanoparticles were used as conductive fillers, while PEO served as the polymeric binder in the water/ethanol-based printable ink formulation. The solid composition consisted of 85 wt% silver flakes, 5 wt% silver nanoparticles, and 10 wt% polyethylene oxide (PEO). A water/ethanol solvent system was used to obtain a printable formulation. The rheological properties of the ink were characterized using a rotational rheometer (MCR301, Anton Paar GmbH, Austria) equipped with a 50-mm-diameter, 1° cone–plate geometry with a gap of 0.101 mm. All rheological measurements were performed at 25 °C. Before the primary measurements, the ink was subjected to a single preshear sweep from 0.01 to 1000 $s^{-1}$. Steady-flow measurements were subsequently performed by decreasing the shear rate from 1000 to 0.01 $s^{-1}$, followed by an increasing sweep from 0.01 to 1000 $s^{-1}$. The resulting apparent viscosity ($\eta$) was recorded as a function of shear rate. Following the steady-flow measurements, strain-amplitude sweeps were performed at a frequency of 1 Hz by increasing the shear strain from 0.01% to 1000%. The storage modulus ($G'$) and loss modulus ($G''$) were recorded as functions of strain to characterize the viscoelastic response of the printable ink.

The conductive silver ink was extruded through a 24-gauge (24G) nozzle using a pneumatic dispensing system. The nozzle was translated at a constant speed of 20 mm $s^{-1}$ during printing. Compressed air at a pressure of 0.2 MPa was supplied to a pneumatic valve controller (PMC-PT-AC, Koganei, Japan), and ink extrusion was controlled by programmatically switching the pneumatic valve on and off in synchronization with the robotic printing trajectory. The compressed-air flow was adjusted to a controller setting of 7% using a flow-control device (IBFL-J6, Koganei, Japan). These printing conditions were used to fabricate the conductive traces evaluated for line width and thickness (Figure 4E,F).

*Fabrication of the insulating layer*

A commercially available pyroxylin-based liquid bandage (Sakamukea, Kobayashi Pharmaceutical Co., Ltd., Japan) was used as the insulating material. The liquid bandage was dispensed directly in its original formulation without dilution using the robotic pneumatic dispensing system. Printing was performed through a 14-gauge (14G) nozzle at a constant nozzle translation speed of 20 mm $s^{-1}$ under an applied pneumatic pressure of 0.2 MPa. The compressed-air flow was adjusted to a controller setting of 8% using a flow-control device (IBFL-J6C, Koganei, Japan). Material dispensing was controlled by programmatically switching the pneumatic valve on and off using a pneumatic valve controller (PMC-PT-AC, Koganei, Japan), synchronized with the robotic printing trajectory. The deposited material was allowed to dry naturally under ambient conditions to form an electrically insulating layer between the printed receiving coil and IDEs.

*Printing characterization and electromechanical evaluation*

The conductive silver ink was robotically printed onto transparent polyurethane dressings to evaluate its suitability for conformal device fabrication. Representative conductive patterns were printed on flexible and curved substrates, and the dimensional reproducibility of the printed traces was evaluated by measuring the line width and printed thickness at multiple locations. The electrical resistance of the printed conductors was monitored immediately after printing to characterize the evolution of conductivity during solvent evaporation and drying. The time-dependent resistance was recorded until the resistance approached a steady value.

The electromechanical stability of the printed conductors was evaluated under deformation representative of conformal use on body surfaces. For cyclic compression testing, the printed samples were subjected to 50% compressive strain for 1000 cycles, while the electrical resistance was periodically monitored. The measured resistance was normalized to the initial resistance ($R_0$) and expressed as $R/R_0$. In a separate tensile test, the printed conductive traces were elongated to strains of up to 50%, and the corresponding change in relative resistance ($R/R_0$) was measured to evaluate electrical continuity under stretching.

*Wireless RF power transfer and electric-field characterization*

Wireless RF signal reception by the curvature- and tissue-aware device was evaluated at the target operating frequency of 25 MHz through magnetic coupling between an external excitation coil and the printed receiving coil. A 25 MHz RF carrier signal with a peak-to-peak voltage of 10 Vpp applied to the external transmitter was pulse-modulated at a repetition frequency of 4 Hz and a duty cycle of 20%, corresponding to an RF-ON duration of 50 ms and an RF-OFF duration of 200 ms per cycle. The pulsed RF excitation induced an RF voltage in the printed receiving structure.

An equivalent circuit was constructed to describe the wireless RF system (Figure 5B). The transmitter side included the electrical characteristics of the external excitation coil, whereas the receiver side incorporated the resistance and inductance of the printed receiving coil and its parasitic capacitance. The load associated with the IDEs and biological interface was represented by the capacitance between the IDEs, the capacitive coupling through the polyurethane dressing, and the resistive and capacitive components of the tissue-equivalent load. The electric field generated by the IDEs was evaluated using an electromagnetic model incorporating the polyurethane dressing and tissue layer. The spatial electric-field distribution was calculated both along the electrode plane and in cross-section through the dressing and underlying tissue. The electric-field magnitude was further evaluated as a function of depth from the tissue surface to quantify attenuation of the delivered field within the tissue-equivalent medium.

The electrical response of the fabricated device was characterized by measuring the impedance magnitude and phase angle as functions of frequency using an impedance analyzer (4294A, Keysight Technologies, USA). The resonant frequency was determined from the frequency-dependent impedance response. To evaluate wireless RF signal transmission, a pulsed RF waveform was applied to the external excitation coil, and the resulting voltage waveforms were measured using a digital oscilloscope (MSO46, Tektronix, USA). The input and received voltage waveforms were compared to confirm wireless transmission of the pulsed RF signal to the printed receiving device.

*Experimental RF validation on an anthropomorphic tissue-equivalent forearm phantom*

An anthropomorphic tissue-equivalent forearm phantom (Access Co., Ltd., Japan) was used to experimentally validate the curvature- and tissue-aware RF design under anatomically relevant conditions. A transparent polyurethane dressing was conformally attached to the phantom surface before robotic fabrication.

The dielectric properties of the phantom measured at 25 MHz were incorporated into the electromagnetic loading model, with a relative permittivity of $\varepsilon_r = 125$ and an electrical conductivity of $\sigma = 0.31\ \mathrm{S\ m^{-1}}$.

The local surface curvature obtained from 3D measurement of the forearm phantom was combined with these experimentally determined dielectric properties to re-optimize the receiving coil and IDE geometry for a target operating frequency of 25 MHz.

To quantify the influence of surface curvature on RF behavior, the frequency-dependent impedance characteristics of the optimized device were numerically compared between planar and curved configurations under the prescribed tissue-equivalent loading conditions. The optimized device was subsequently fabricated directly on the transparent dressing attached to the anthropomorphic forearm phantom using the 6-DOF robotic system. Following fabrication, the impedance magnitude and phase angle were experimentally measured as functions of frequency. The measured resonant frequency was compared with the corresponding simulation result and the target frequency of 25 MHz to evaluate the accuracy of the curvature- and tissue-aware RF design under anatomically conformal conditions (Figure 6).

**Acknowledgments**

The authors thank Mingeon Kim for technical assistance with the robotic control software and support during the experiments.

**Author Contributions**

H.T. conceived and initiated the research project, designed and conducted the experiments, developed the robotic control software, performed data analysis, and wrote the original manuscript. J.M. fabricated and characterized the wireless system and performed the electrical measurements. N.K. performed the rheological characterization of the conductive silver ink. M.H. reviewed and edited the manuscript. All authors reviewed and approved the final version of the manuscript.

**References**

1 Zhu, Z. *et al.* 3D Printed Functional and Biological Materials on Moving Freeform Surfaces. *Adv Mater* **30**, e1707495 (2018). https://doi.org/10.1002/adma.201707495

2 Ma, K. *et al.* Application of robotic-assisted in situ 3D printing in cartilage regeneration with HAMA hydrogel: An in vivo study. *J Adv Res* **23**, 123-132 (2020). https://doi.org/10.1016/j.jare.2020.01.010

3 Li, L. *et al.* Robotic in situ 3D bio-printing technology for repairing large segmental bone defects. *J Adv Res* **30**, 75-84 (2021). https://doi.org/10.1016/j.jare.2020.11.011

4 Zhou, C. *et al.* Ferromagnetic soft catheter robots for minimally invasive bioprinting. *Nat Commun* **12**, 5072 (2021). https://doi.org/10.1038/s41467-021-25386-w
5 Zhao, W. *et al.* Subaqueous Bioprinting: A Novel Strategy for Fetal Membrane Repair with 7-Axis Robot-Assisted Minimally Invasive Surgery. *Advanced Functional Materials* **32** (2022). https://doi.org/10.1002/adfm.202207496
6 Kim, J. *et al.* Large Scale Ultrafast Manufacturing of Wireless Soft Bioelectronics Enabled by Autonomous Robot Arm Printing Assisted by a Computer Vision-Enabled Guidance System for Personalized Wound Healing. *Adv Healthc Mater* **14**, e2401735 (2025). https://doi.org/10.1002/adhm.202401735
7 Zhu, Z., Ng, D. W. H., Park, H. S. & McAlpine, M. C. 3D-printed multifunctional materials enabled by artificial-intelligence-assisted fabrication technologies. *Nature Reviews Materials* **6**, 27-47 (2020). https://doi.org/10.1038/s41578-020-00235-2
8 Hakimi, N. *et al.* Handheld skin printer: in situ formation of planar biomaterials and tissues. *Lab Chip* **18**, 1440-1451 (2018). https://doi.org/10.1039/c7lc01236e
9 Urciuolo, A. *et al.* Intravital three-dimensional bioprinting. *Nat Biomed Eng* **4**, 901-915 (2020). https://doi.org/10.1038/s41551-020-0568-z
10 Zhao, M. Electrical fields in wound healing-An overriding signal that directs cell migration. *Semin Cell Dev Biol* **20**, 674-682 (2009). https://doi.org/10.1016/j.semcdb.2008.12.009
11 Rodrigues, M., Kosaric, N., Bonham, C. A. & Gurtner, G. C. Wound Healing: A Cellular Perspective. *Physiol Rev* **99**, 665-706 (2019). https://doi.org/10.1152/physrev.00067.2017
12 Derakhshandeh, H., Kashaf, S. S., Aghabaglou, F., Ghanavati, I. O. & Tamayol, A. Smart Bandages: The Future of Wound Care. *Trends Biotechnol* **36**, 1259-1274 (2018). https://doi.org/10.1016/j.tibtech.2018.07.007
13 Jeong, S.-H. *et al.* Accelerated wound healing with an ionic patch assisted by a triboelectric nanogenerator. *Nano Energy* **79** (2021). https://doi.org/10.1016/j.nanoen.2020.105463
14 Jiang, Y. *et al.* Wireless, closed-loop, smart bandage with integrated sensors and stimulators for advanced wound care and accelerated healing. *Nat Biotechnol* **41**, 652-662 (2023). https://doi.org/10.1038/s41587-022-01528-3
15 Shirzaei Sani, E. *et al.* A stretchable wireless wearable bioelectronic system for multiplexed monitoring and combination treatment of infected chronic wounds. *Sci Adv* **9**, eadf7388 (2023). https://doi.org/10.1126/sciadv.adf7388
16 He, Z. *et al.* Highly stretchable, deformation-stable wireless powering antenna for wearable electronics. *Nano Energy* **112** (2023). https://doi.org/10.1016/j.nanoen.2023.108461
17 Gabriel, S., Lau, R. W. & Gabriel, C. The dielectric properties of biological tissues: III. Parametric models for the dielectric spectrum of tissues. *Phys Med Biol* **41**, 2271-2293 (1996). https://doi.org/10.1088/0031-9155/41/11/003
18 Chen, Y. *et al.* Noninvasive in vivo 3D bioprinting. *Sci Adv* **6**, eaba7406 (2020). https://doi.org/10.1126/sciadv.aba7406
19 Mengaldo, G. *et al.* A concise guide to modelling the physics of embodied intelligence in soft robotics. *Nature Reviews Physics* **4**, 595-610 (2022). https://doi.org/10.1038/s42254-022-00481-z
20 Xiang, K. *et al.* Aligning Perception, Reasoning, Modeling and Interaction: A Survey on Physical AI. *IEEE Trans Pattern Anal Mach Intell* **PP** (2026). https://doi.org/10.1109/TPAMI.2026.3703038
21 Yoo, C. *et al.* Robotic Arm-Assisted Conformal 3D Printing of Displays for Structural Electronics. *Advanced Intelligent Systems* **8** (2026). https://doi.org/10.1002/aisy.202501340

22 Meng, Y., Ma, T., Pavinatto, F. J. & MacKenzie, J. D. Interface Modified Flexible Printed Conductive Films via Ag(2)O Nanoparticle Decorated Ag Flake Inks. *ACS Appl Mater Interfaces* **11**, 9190-9196 (2019). https://doi.org/10.1021/acsami.8b20057
23 You, F. *et al.* Research on the modification of ink by introducing silver and SiO2@Ag composite nanoparticles. *Materials Research Express* **12** (2025). https://doi.org/10.1088/2053-1591/ae2786
24 Inoue, M. & Liu, J. Effects of Multi-modal Filler Size Distributions on Thermal Conductivity of Electrically Conductive Adhesives Containing Ag Micro and Nanoparticles. *Transactions of The Japan Institute of Electronics Packaging* **2**, 125-133 (2009). https://doi.org/10.5104/jiepeng.2.125

# Supplementary Materials for

## AI–guided robotic in situ fabrication of curvature- and tissue-adaptive wireless bioelectronics

*Hiroyuki Tetsuka*, Jiaju Ma, Naomi Kumano, Minoru Hirano*

Toyota Central R&D Labs., Inc., 41-1 Yokomichi, Nagakute, Aichi 480-1192, Japan
E-mail: h-tetsuka@mosk.tytlabs.co.jp

### Supplementary Text S1: Eye-in-hand calibration between the robotic arm and RGB-D camera

To establish a geometric transformation between the RGB-D camera coordinate system and the robot tool coordinate system, a six-degree-of-freedom eye-in-hand calibration was performed using a ChArUco calibration target. The camera was mounted on the robotic manipulator, whereas the ChArUco target remained stationary within the robot workspace. During calibration, the robot was sequentially moved through a predefined set of translational and rotational poses to provide sufficient diversity in both position and orientation.

A ChArUco board based on the DICT_5X5_1000 ArUco dictionary was used as the calibration target. The board configuration was defined as 13 × 9, with a chessboard square size of 30 mm and an ArUco marker size of 21 mm. Images were acquired using an Orbbec RGB-D camera. In the default RGB acquisition configuration, color images were acquired at 1920 × 1080 pixels and 30 frames $s^{-1}$, while depth images were acquired at 640 × 576 pixels and 30 frames $s^{-1}$.

### Coordinate-system definition

Throughout the calibration procedure, a homogeneous transformation matrix was defined as the transformation that maps coordinates expressed in frame B to frame A.

$$ {}^{\mathrm{A}}T^{\mathrm{B}} = \begin{bmatrix} {}^{\mathrm{A}}R^{\mathrm{B}} & {}^{\mathrm{A}}t^{\mathrm{B}} \\ 0^{\mathrm{T}} & 1 \end{bmatrix} $$

The robot controller provided the TCP pose with respect to the robot base frame, denoted ${}^{\mathrm{B}}T_i^{\mathrm{G}}$, where B denotes the robot base frame, G denotes the TCP (gripper/tool) frame, and $i$ denotes the calibration sample. The translational components reported by the robot controller were converted from millimeters to meters.

The orientation values $(U, V, W)$ were converted to a rotation matrix according to the convention used in the implementation:

$$ {}^{\mathrm{B}}R^{\mathrm{G}} = R_z(U)\, R_y(V)\, R_x(W). $$

For each robot pose, ChArUco corners were detected in the camera image and the pose of the target relative to the camera was estimated using OpenCV. This provided ${}^{\mathrm{C}}T_i^{\mathrm{T}}$, where C denotes the camera frame and T denotes the ChArUco target frame. Therefore, a three-dimensional point expressed in the target coordinate system was related to the camera coordinate system as

$$ p^{\mathrm{C}} = {}^{\mathrm{C}}R^{\mathrm{T}}\, p^{\mathrm{T}} + {}^{\mathrm{C}}t^{\mathrm{T}}. $$

The target-frame origin was defined at the center of the ChArUco pattern rather than at the default corner-based origin used by OpenCV. If $c^{\mathrm{T}}$ denotes the offset between the original board origin and the board center, the corrected translation was calculated as

$$ {}^{\mathrm{C}}t^{\mathrm{T}}_{\mathrm{center}} = {}^{\mathrm{C}}t^{\mathrm{T}}_{\mathrm{corner}} + {}^{\mathrm{C}}R^{\mathrm{T}}\, c^{\mathrm{T}}. $$

**Synchronized robot-pose and image acquisition**

The robot was moved sequentially through a predefined pose sequence containing variations in Cartesian position and orientation. After completion of each robot motion, an additional settling period was introduced to minimize temporal mismatch between the robot state and the camera measurement. Buffered camera frames were subsequently discarded before acquisition of the calibration image to reduce the possibility of associating a stale image with the current robot pose.

Immediately before accepting a calibration sample, the measured robot pose was compared with the commanded pose. A sample was accepted only when the maximum Cartesian position error was less than or equal to 1.0 mm and the maximum orientation error was less than or equal to 0.5°. The robot state was queried up to five times when these criteria were not initially satisfied. Samples that failed this consistency check were discarded.

For each accepted pose, the measured robot transformation ${}^{\mathrm{B}}T_i^{\mathrm{G}}$ and the corresponding vision-based transformation ${}^{\mathrm{C}}T_i^{\mathrm{T}}$ were stored as a synchronized calibration pair.

**ChArUco detection quality assessment**

To reduce the influence of poorly detected calibration images, each ChArUco observation was characterized by the number of detected ChArUco corners, the reprojection error, and the image coverage of the detected target.

The reprojection root-mean-square error was calculated as

$$e_{\mathrm{rep},i} = \sqrt{\frac{1}{N_i}\sum_{j=1}^{N_i} \left\| u_{ij}^{(\mathrm{obs})} - u_{ij}^{(\mathrm{proj})} \right\|_2^2}.$$

Here, $N_i$ is the number of detected ChArUco corners in image $i$, $u_{ij}^{(\mathrm{obs})}$ is the measured pixel location of corner $j$, and $u_{ij}^{(\mathrm{proj})}$ is its pixel location obtained by reprojecting the estimated three-dimensional board pose using the camera intrinsic parameters.

Spatial coverage was quantified from the bounding box containing all detected ChArUco corners:

$$C_i = \frac{A_{\mathrm{bbox},i}}{A_{\mathrm{image}}}.$$

When quality filtering was enabled, samples were first rejected according to thresholds for the minimum number of detected ChArUco corners, maximum reprojection RMSE, and minimum target coverage. The implementation additionally supported a second robust filtering stage based on the median absolute deviation (MAD):

$$\mathrm{MAD}(x) = \mathrm{median}(|x - \mathrm{median}(x)|).$$

Samples with abnormally large reprojection errors or abnormally low target coverage relative to the corresponding median and MAD values were excluded prior to hand-eye calibration.

**Eye-in-hand transformation estimation**

The unknown eye-in-hand transformation was defined as

$$X = {}^{\mathrm{G}}T^{\mathrm{C}},$$

representing the transformation from the camera coordinate system to the robot TCP coordinate system. Because the ChArUco board remained stationary with respect to the robot base, the estimated transformation had to satisfy

$${}^{\mathrm{B}}T^{\mathrm{T}} = {}^{\mathrm{B}}T_i^{\mathrm{G}}\ {}^{\mathrm{G}}T^{\mathrm{C}}\ {}^{\mathrm{C}}T_i^{\mathrm{T}}$$

for every calibration sample $i$. Therefore,

$${}^{\mathrm{B}}T_i^{\mathrm{G}}\, X\ {}^{\mathrm{C}}T_i^{\mathrm{T}} = \text{constant}.$$

For two calibration poses $i$ and $j$, this relationship can be rearranged into the conventional hand-eye calibration form

$$A_{ij}\, X = X\, B_{ij},$$

where $A_{ij}$ and $B_{ij}$ describe the relative robot and target motions, respectively. The transformation $X$ was estimated using the OpenCV calibrateHandEye implementation. Five established hand-eye calibration algorithms were evaluated independently: Tsai-Lenz, Park-Martin, Horaud, Daniilidis, and Andreff.

**Bootstrap-based selection of the calibration algorithm**

The stability of each hand-eye calibration algorithm was evaluated using random subsampling. First, the transformation obtained using the complete calibration dataset was taken as the reference result for each algorithm. Subsequently, 200 calibration calculations were performed using randomly selected subsets containing 70% of the available samples.

For each bootstrap calculation $k$, the translational deviation from the full-data solution was calculated as

$$e_{t,k} = 1000\|t_k - t_{\mathrm{ref}}\|_2 \quad [\mathrm{mm}],$$

whereas the rotational deviation was calculated from

$$\Delta R_k = R_{\mathrm{ref}}^{\mathrm{T}} R_k$$

as

$$e_{r,k} = \cos^{-1}\left(\frac{\mathrm{tr}(\Delta R_k) - 1}{2}\right) \quad [\mathrm{deg}].$$

The 95th-percentile translational and rotational deviations, denoted $t_{95}$ and $r_{95}$, respectively, were combined into a stability score:

$$S = t_{95} + 0.5\, r_{95}.$$

The algorithm yielding the minimum value of $S$ was selected as the final hand-eye calibration method for the dataset.

**Residual-based refinement**

Following the initial hand-eye calibration, an additional consistency analysis was performed using the stationary-target constraint. For each sample,

$${}^{\mathrm{B}}T_i^{\mathrm{T}} = {}^{\mathrm{B}}T_i^{\mathrm{G}}\ {}^{\mathrm{G}}T^{\mathrm{C}}\ {}^{\mathrm{C}}T_i^{\mathrm{T}}$$

was calculated. Because the physical target was stationary, all estimates of ${}^{\mathrm{B}}T_i^{\mathrm{T}}$ should ideally be identical.

A representative target transformation, ${}^{\mathrm{B}}\overline{T}^{\mathrm{T}}$, was obtained using robust averaging, in which rotations were averaged in quaternion space and translations were combined using their median values. The residual transformation for sample $i$ was then defined as

$$E_i = \left({}^{\mathrm{B}}\overline{T}^{\mathrm{T}}\right)^{-1}\ {}^{\mathrm{B}}T_i^{\mathrm{T}}.$$

The translational residual was

$$e_{t,i}^{(\mathrm{res})} = 1000\|E_i(1:3,4)\|_2 \quad [\mathrm{mm}],$$

and the rotational residual was

$$e_{r,i}^{(\mathrm{res})} = \cos^{-1}\left(\frac{\mathrm{tr}\big(E_i(1:3,1:3)\big) - 1}{2}\right) \quad [\mathrm{deg}].$$

To identify samples belonging to the high-error tail of the residual distribution, the translation and rotation residuals were independently normalized using robust median-based scores. A combined residual score was calculated as

$$s_i = \left|\frac{e_{t,i}^{(\mathrm{res})} - \mathrm{median}\left(e_t^{(\mathrm{res})}\right)}{\mathrm{MAD}\left(e_t^{(\mathrm{res})}\right)}\right| + \left|\frac{e_{r,i}^{(\mathrm{res})} - \mathrm{median}\left(e_r^{(\mathrm{res})}\right)}{\mathrm{MAD}\left(e_r^{(\mathrm{res})}\right)}\right|.$$

Samples satisfying

$$s_i > \mathrm{median}(s) + 3.5\,\mathrm{MAD}(s)$$

were considered candidate residual outliers, provided that a sufficiently large tail population was detected. The hand-eye transformation was then recalculated after removal of the selected samples.

To prevent over-filtering, the refined solution was retained only when it improved the combined 95th-percentile residual metric

$$S_{\mathrm{res}} = e_{t,95}^{(\mathrm{res})} + 0.5\, e_{r,95}^{(\mathrm{res})}$$

without increasing the sample-to-sample translational variability of the reconstructed stationary target. Otherwise, the algorithm automatically reverted to the initial calibration solution.

### Transformation to the depth-camera coordinate system

When the calibration was performed using the RGB camera, the RGB-to-depth extrinsic transformation provided by the Orbbec camera calibration was used to express the result in the depth-camera coordinate system. If ${}^{\mathrm{C}}T^{\mathrm{D}}$ denotes the transformation from the depth-camera coordinate system to the color-camera coordinate system, the final depth-camera-to-TCP transformation was calculated as

$${}^{\mathrm{G}}T^{\mathrm{D}} = {}^{\mathrm{G}}T^{\mathrm{C}}\ {}^{\mathrm{C}}T^{\mathrm{D}}.$$

The resulting 4 × 4 homogeneous transformation matrix was stored and subsequently used to transform three-dimensional points reconstructed from depth measurements into the robot coordinate system for robot-guided conformal printing.

### Implementation parameters

| Parameter | Implementation | Purpose |
|---|---|---|
| Calibration target | ChArUco, DICT_5X5_1000; 13 × 9; 30-mm squares; 21-mm markers | Target pose estimation |
| Robot-pose acceptance | ≤ 1.0 mm position error; ≤ 0.5° orientation error; up to 5 retries | Reject pose/image mismatches |
| Hand-eye solvers | Tsai-Lenz, Park-Martin, Horaud, Daniilidis, Andreff | Algorithm comparison |
| Bootstrap evaluation | 200 repetitions; 70% subsampling | Estimate calibration stability |
| Solver score | S = t95 + 0.5 r95 | Select the most stable solver |
| Residual tail threshold | median + 3.5 × MAD | Candidate outlier detection |
| Output | 4 × 4 camera-to-TCP transform; optional depth-camera transform | Robot/camera coordinate registration |

## Supplementary Text S2: Robotic pose generation and control for conformal coil printing

### Robotic trajectory generation on the reconstructed wound surface

After generation of the planar coil centerline, the printing trajectory was mapped onto the three-dimensional wound surface reconstructed as a triangular mesh. The coil pattern was centered with respect to the detected wound region, and the pattern coordinates were transformed from the local two-dimensional $(u, v)$ coordinate system to the three-dimensional robot base coordinate system. The resulting points were then projected onto the mesh surface along the estimated local surface normal. When ray casting failed for individual points, nearest-neighbor snapping to the sampled mesh surface was used as a fallback to preserve trajectory continuity.

### Surface-normal-based tool orientation

For each projected surface point $p_i \in \mathrm{R}^3$, the local surface normal obtained from the mesh was used to determine the orientation of the printing tool. To avoid abrupt inversion of the normal direction, the normal was first aligned with the $z$-axis of the reference tool orientation and was subsequently constrained to remain directionally consistent with the normal at the preceding trajectory point. The normal was then low-pass filtered according to

$$n_i = \frac{(1-\alpha)n_{i-1} + \alpha\tilde{n}_i}{\|(1-\alpha)n_{i-1} + \alpha\tilde{n}_i\|},$$

where $\tilde{n}_i$ is the raw mesh normal at the $i$-th point and $\alpha = 0.2$ is the normal-smoothing coefficient used in the present implementation.

The local tangential direction of the printing path was calculated from the spatial derivative of consecutive trajectory points. Let $\Delta p_i$ denote the local trajectory direction estimated numerically by finite differences. Its component parallel to the local tangent plane was calculated as

$$t_i = \frac{(I - n_i n_i^{\mathrm{T}})\Delta p_i}{\|(I - n_i n_i^{\mathrm{T}})\Delta p_i\|}.$$

A reference tool $x$-axis, $x_{\mathrm{ref}}$, obtained from the initial robot pose was also projected onto the current tangent plane:

$$x_{\mathrm{ref},i} = \frac{(I - n_i n_i^{\mathrm{T}})x_{\mathrm{ref}}}{\|(I - n_i n_i^{\mathrm{T}})x_{\mathrm{ref}}\|}.$$

The desired in-plane tool direction was then constructed as a weighted combination of the projected reference direction and the local trajectory tangent:

$$x_{\mathrm{seed},i} = \frac{\beta x_{\mathrm{ref},i} + (1-\beta)t_i}{\|\beta x_{\mathrm{ref},i} + (1-\beta)t_i\|},$$

with $\beta = 0.5$. To further improve orientation continuity, the $x$-axis of the preceding tool pose was projected onto the current tangent plane and blended with $x_{\mathrm{seed},i}$. The sign of the resulting axis was selected to prevent 180° flips between consecutive trajectory points.

An orthonormal local tool coordinate system was subsequently constructed as

$$z_i = n_i,$$

$$y_i = \frac{z_i \times x_{\mathrm{seed},i}}{\|z_i \times x_{\mathrm{seed},i}\|},$$

$$x_i = \frac{y_i \times z_i}{\|y_i \times z_i\|},$$

and the desired end-effector rotation matrix was expressed as

$$R_i = [x_i \quad y_i \quad z_i].$$

This formulation maintained the printing-tool $z$-axis approximately normal to the local wound surface while simultaneously providing a continuous in-plane orientation along the printing trajectory.

### Orientation continuity and angular-rate limiting

To suppress sudden rotational motion caused by local mesh irregularities, the angular change between consecutive tool orientations was explicitly limited. The relative rotation between consecutive poses was calculated as

$$R_{\mathrm{rel},i} = R_{i-1}^{\mathrm{T}} R_i.$$

Using the axis-angle representation, the corresponding rotation vector $r_i$ and angular magnitude $\theta_i$ were obtained from

$$r_i = \log(R_{\mathrm{rel},i}), \qquad \theta_i = \|r_i\|.$$

The maximum permitted orientation change was made proportional to the traveled distance:

$$\theta_{\mathrm{max},i} = k_\theta \Delta s_i,$$

where $\Delta s_i$ is the displacement between consecutive trajectory points and $k_\theta = 0.25^\circ\ \mathrm{mm}^{-1}$ in the present implementation. When $\theta_i$ exceeded $\theta_{\mathrm{max},i}$, the desired orientation was limited according to

$$R_i^* = R_{i-1} \exp\left[\frac{\theta_{\mathrm{max},i}}{\theta_i}[r_i]_\times\right],$$

where $[r_i]_\times$ denotes the skew-symmetric matrix associated with the rotation vector. This constraint reduced abrupt tool reorientation over locally irregular surface regions.

### TCP positioning and trajectory resampling

The tool-center-point (TCP) position was defined from the projected surface point and the local tool $z$-axis as

$$p_{\mathrm{TCP},i} = p_i + d_\mathrm{n} z_i,$$

where $d_\mathrm{n}$ is the prescribed normal offset between the reconstructed surface and the TCP. In the present implementation, $d_\mathrm{n} = -1$ mm, with the sign defined relative to the calculated tool $z$-axis. Thus, both the TCP position and orientation were determined directly from the local surface geometry.

The trajectory was subsequently resampled at an approximately uniform arc-length interval of 1 mm. Cartesian positions were linearly interpolated, whereas orientations were interpolated in quaternion space using spherical linear interpolation (SLERP). The interpolated rotation matrices were converted to the Euler-angle representation required by the robot controller. In the present implementation, the orientation convention was

$$R = \mathrm{R}_z(U)\,\mathrm{R}_y(V)\,\mathrm{R}_x(W).$$

Angle unwrapping was applied to suppress discontinuities associated with the 360° periodicity of Euler angles.

**Robot control and path execution**

Each resampled trajectory point was represented by the six-dimensional Cartesian robot command

$$q_i = [X_i \quad Y_i \quad Z_i \quad U_i \quad V_i \quad W_i]^\mathrm{T},$$

where $(X_i, Y_i, Z_i)$ specify the desired TCP position in the robot base coordinate system and $(U_i, V_i, W_i)$ specify its orientation. The flange-to-TCP transformation was defined in the Epson RC+ tool configuration so that the calculated Cartesian coordinates could be directly interpreted as TCP commands by the robot controller.

The trajectory was transmitted to the Epson robot controller through a TCP/IP interface and executed using Cartesian Move XY commands. Continuous-path motion was enabled between successive intermediate points to obtain smooth motion along the curved surface. Before and after each printing trajectory, the tool was retracted by 30 mm in the robot $Z$ direction to provide a collision-safe approach and withdrawal motion.

For the integrated coil-IDE device, the coil, insulating layer, and the two IDE electrodes were treated as independent robot paths. The robot therefore executed the coil trajectory first, followed by a retraction motion and separate trajectories for insulating layer, IDE-A, and IDE-

B. This four-pass procedure provided independent control of the conductive printing paths and reduced the risk of unintended connection between electrically distinct structures.

## Implementation parameters

| Parameter | Value | Role |
| --- | --- | --- |
| Normal smoothing coefficient, $\alpha$ | 0.2 | Low-pass filtering of mesh normals |
| Reference/tangent weighting, $\beta$ | 0.5 | In-plane tool-axis construction |
| Maximum orientation change, $k\theta$ | $0.25° \text{ mm}^{-1}$ | Limits pose change per traveled distance |
| Normal TCP offset, d_n | −1 mm | TCP displacement relative to local surface normal |
| Trajectory resampling interval | 1 mm | Approximately uniform arc-length spacing |
| Approach/withdrawal distance | 30 mm | Collision-safe motion before and after printing |

## Supplementary Figures

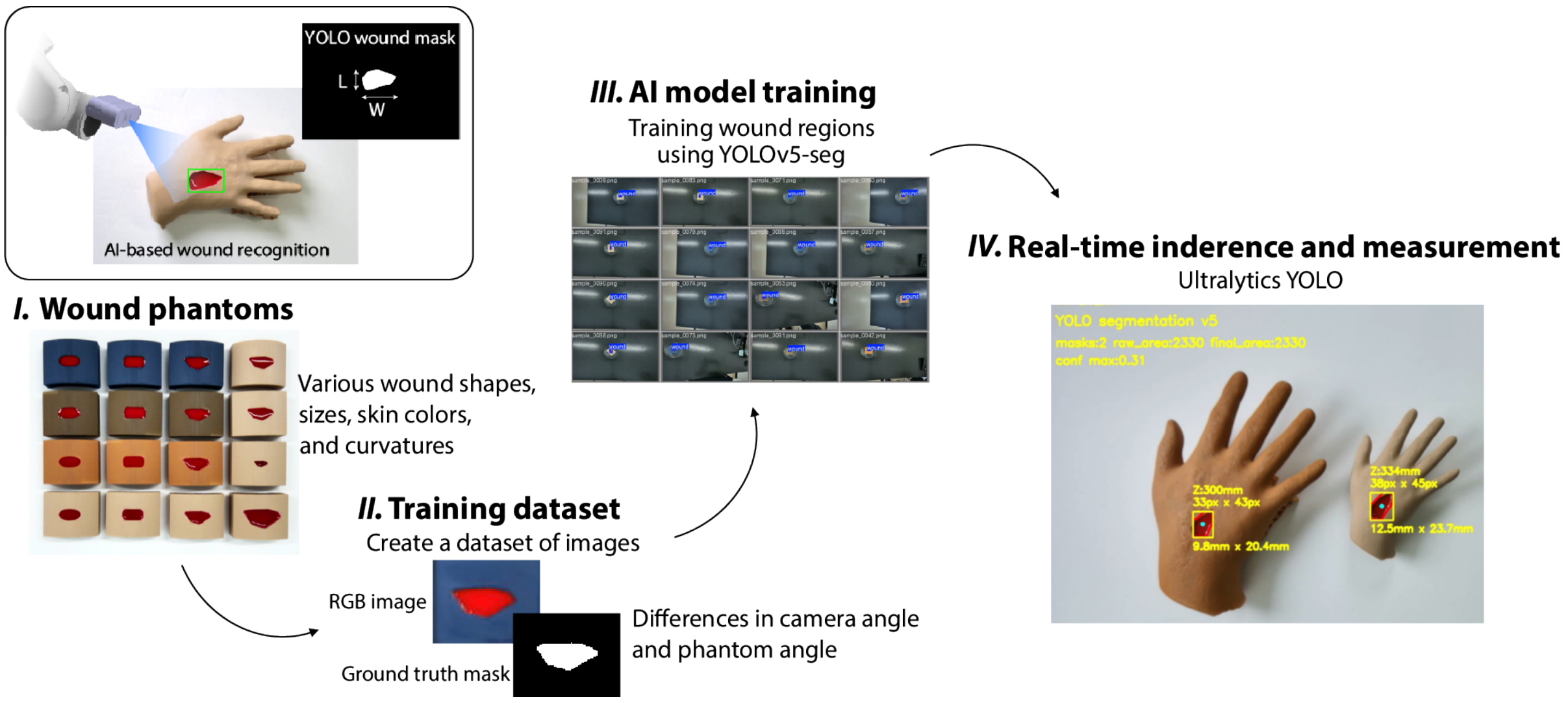

**Figure S1. Workflow for AI-based wound segmentation and real-time geometrical measurement.** (I) Wound phantoms with different wound shapes, sizes, skin colors, and surface curvatures used for model development and evaluation. (II) Preparation of the training dataset consisting of paired RGB images and corresponding ground-truth wound masks acquired under different camera and phantom orientations. (III) Training of the wound-segmentation model using YOLO-seg to identify wound regions at the pixel level. (IV) Representative real-time inference and measurement using the trained Ultralytics YOLO model. The detected wound masks were used to determine wound dimensions for subsequent geometrical characterization and personalized device design.

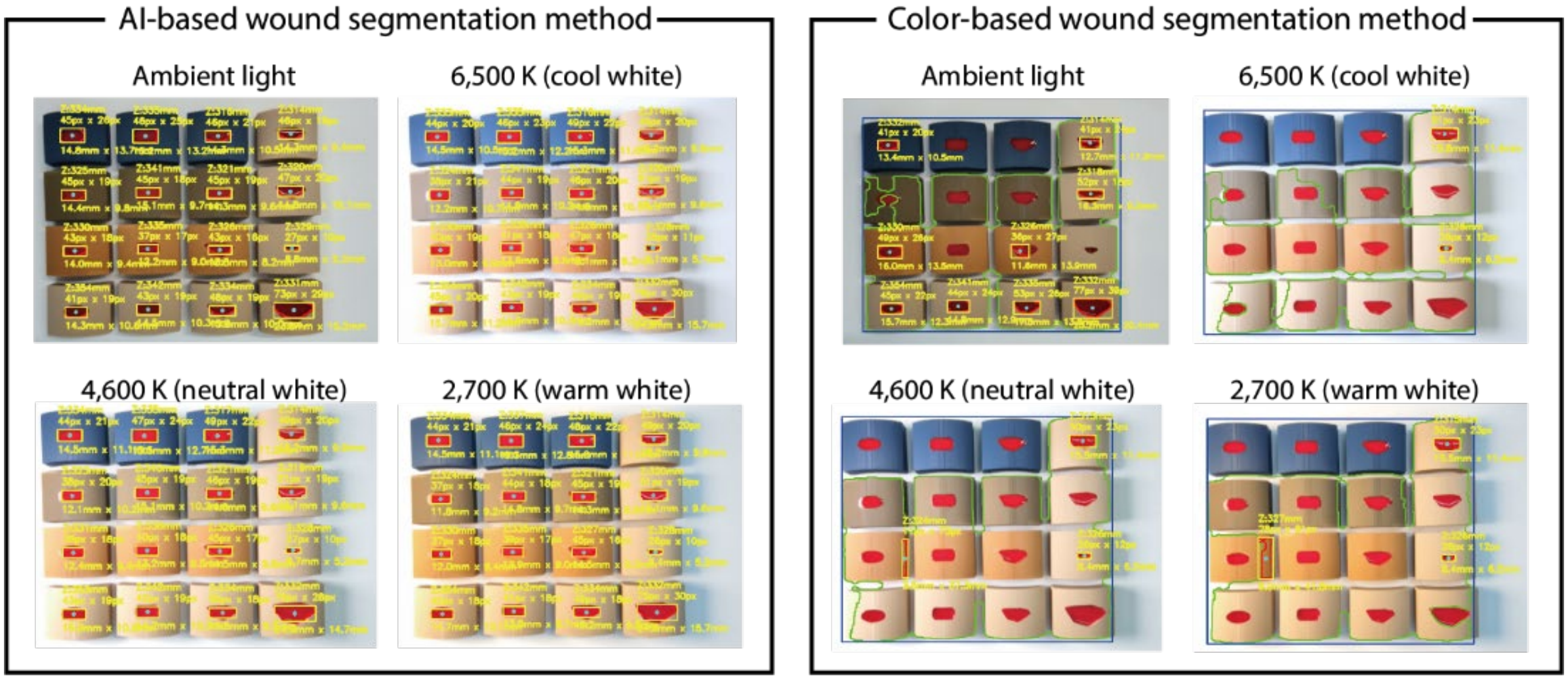

Detection rate

| | Ambient light | 6,500 K light | 4,600 K light | 2,700 K light |
|---|---|---|---|---|
| AI-based method | 16/16 | 16/16 | 16/16 | 16/16 |
| Color-based method | 9/16 | 2/16 | 2/16 | 2/16 |

**Figure S2. Robustness of AI-based wound segmentation under different illumination conditions.** Representative wound-detection results obtained using the AI-based wound segmentation method (left) and a conventional color-based wound segmentation method (right) under four illumination conditions: ambient light, 6,500 K (cool white), 4,600 K (neutral white), and 2,700 K (warm white). Sixteen wound phantoms with different wound geometries and skin colors were evaluated under each illumination condition. The AI-based method successfully detected all 16 wounds under all tested illumination conditions (16/16), whereas the color-based method showed substantially lower detection rates of 9/16 under ambient light and 2/16 under each controlled color-temperature condition. These results demonstrate the robustness of the AI-based segmentation approach against variations in illumination color temperature compared with conventional color-based segmentation.

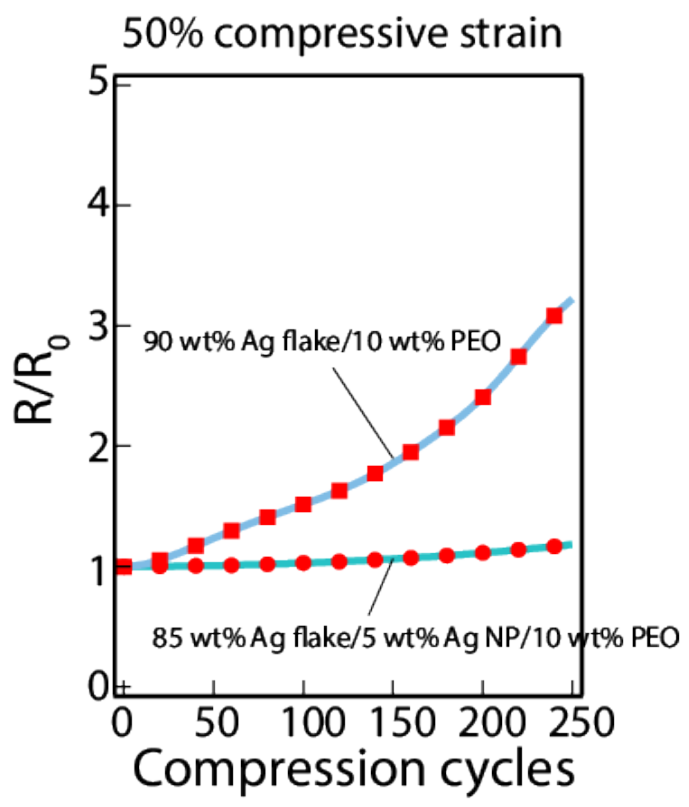

**Figure S3.** Relative resistance ($R/R_0$) of printed conductors with different Ag filler compositions during 250 compression cycles at 50% compressive strain. The conductor containing 85 wt% Ag flakes, 5 wt% Ag nanoparticles, and 10 wt% PEO exhibited a smaller increase in relative resistance than that containing 90 wt% Ag flakes and 10 wt% PEO, indicating improved electrical stability under repeated compression.

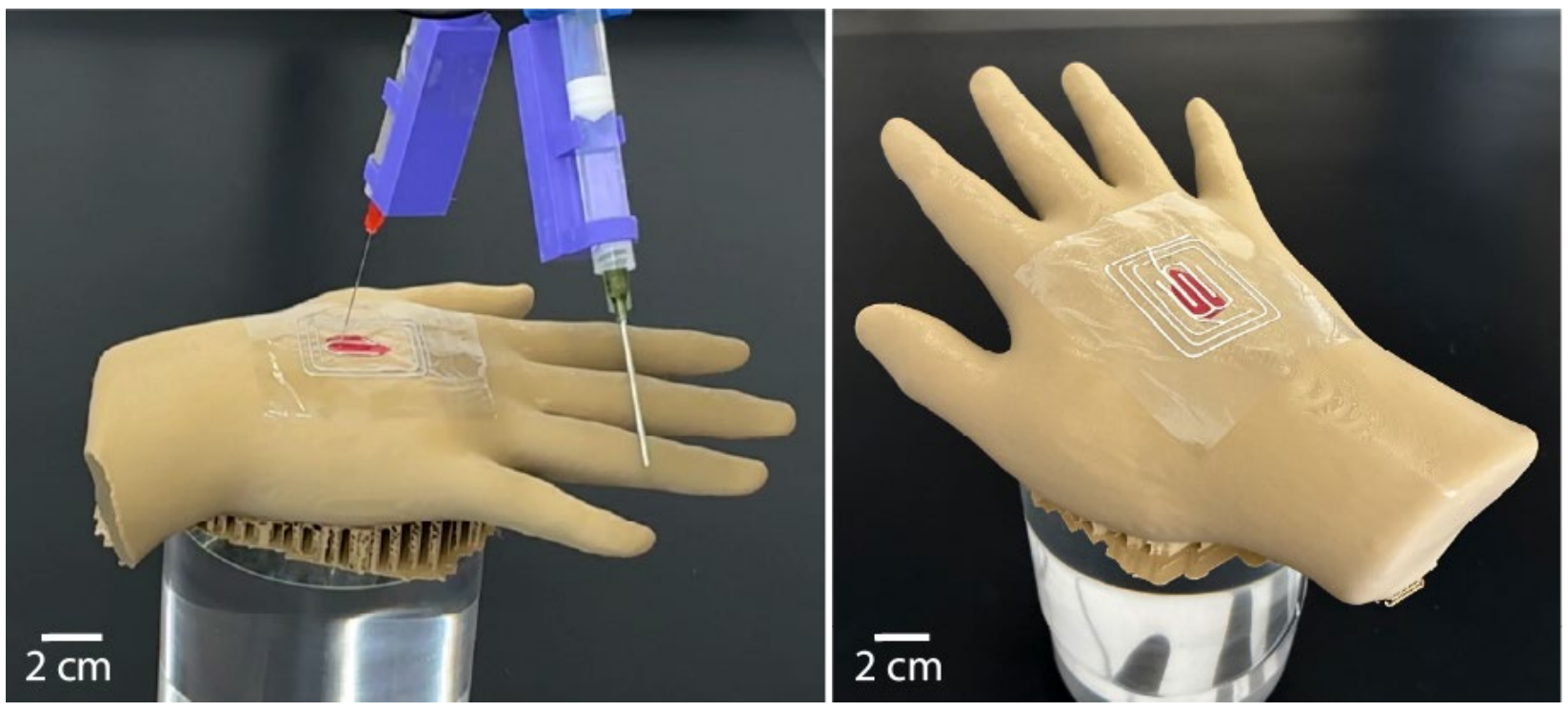

**Figure S4.** Representative photographs of robotic in situ printing of the wireless electrotherapy device onto a transparent dressing conformally attached to a hand-shaped phantom and the resulting printed device.

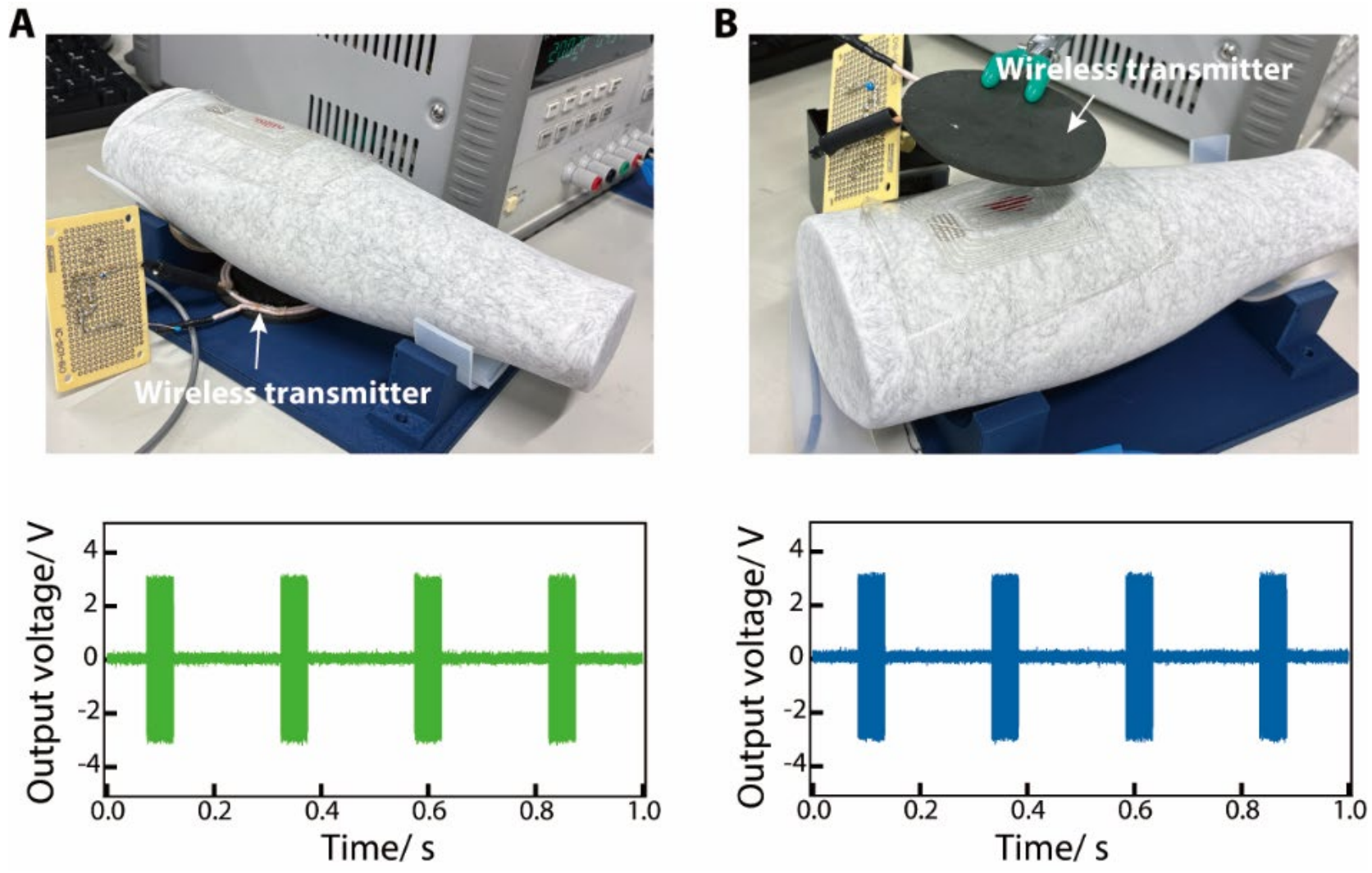

**Figure S5. Wireless RF signal reception under different transmitter configurations on an anthropomorphic tissue-equivalent forearm phantom.** (A, B) Representative photographs of two wireless transmitter configurations relative to the printed receiving device on the tissue-equivalent forearm phantom (top), together with the corresponding measured output voltage waveforms (bottom). (A) Wireless transmitter positioned beneath the forearm phantom, with a transmitter-to-receiver separation distance of 7 cm. (B) Wireless transmitter positioned above the printed device, with a transmitter-to-receiver separation distance of 4 cm. Pulsed RF signals were successfully received under both configurations, demonstrating wireless signal transfer to the printed bioelectronic device on the curved tissue-equivalent substrate.